\documentclass[10pt,twocolumn,letterpaper]{article}

\usepackage{wacv}              

\usepackage{booktabs}
\usepackage[table]{xcolor}
\usepackage{graphicx}
\usepackage{lineno}
\usepackage{subcaption}
\usepackage{caption}
\definecolor{Gray}{gray}{0.9}
\definecolor{gaincolor}{RGB}{232,244,255}
\usepackage{lineno}
\usepackage{float}

\usepackage[ruled,vlined]{algorithm2e}
\usepackage{color}
\usepackage{colortbl}
\usepackage{xcolor}
\usepackage{tikz}
\usepackage{comment}
\usepackage{bibunits}

\definecolor{gold}{HTML}{BD820B}
\definecolor{silver}{HTML}{909090}
\definecolor{bronze}{HTML}{9A5F26}
\definecolor{lgray}{gray}{0.95}

\definecolor{gaincolor}{RGB}{230, 245, 255}

\definecolor{Gray}{gray}{0.91}
\definecolor{LightCyan}{rgb}{0.82,0.82,1}
\newcolumntype{a}{>{\columncolor{Gray}}c}
\newcolumntype{B}{>{\columncolor{LightCyan}}c}

\definecolor{dullgreen}{RGB}{40,120,60}
\definecolor{dullyellow}{RGB}{180,150,40}
\definecolor{dullred}{RGB}{170,60,60}
\definecolor{dullblue}{RGB}{60,90,160}
\definecolor{dullcyan}{RGB}{80,140,170}
\definecolor{wacvblue}{rgb}{0.21,0.49,0.74}
\usepackage[pagebackref,breaklinks,colorlinks,allcolors=wacvblue]{hyperref}

\usepackage{enumitem}
\usepackage[toc,page]{appendix}
\usepackage{titletoc}

\def\confName{WACV}
\def\confYear{2027}

\title{E-MAPS: Event-Based Promptable Video Segmentation with Memory-as-Prompt}

\author{
\mbox{
Adonay Demewez Gebremedhin\textsuperscript{1} \quad
Mohamad Alansari\textsuperscript{2} \quad
Wessam Shehieb\textsuperscript{1} \quad
Muzammal Naseer\textsuperscript{2,3}
}\\
\mbox{
Hasan AlMarzouqi\textsuperscript{2} \quad
Sajid Javed\textsuperscript{2} \quad
Naoufel Werghi\textsuperscript{2}
}\\[0.6em]
\mbox{
\textsuperscript{1}Ajman University, UAE
\qquad
\textsuperscript{2}Khalifa University, UAE
}\\
\textsuperscript{3}University of Western Australia, Australia
\\[0.5em]
\shortstack[c]{
{\scriptsize\ttfamily
202311465@ajmanuni.ac.ae \quad
100061914@ku.ac.ae \quad
w.shehieb@ajman.ac.ae \quad
muhammadmuzammal.naseer@ku.ac.ae
}\\[3pt]
{\scriptsize\ttfamily
\{hasan.almarzouqi,sajid.javed,Naoufel.Werghi\}@ku.ac.ae
}
}
}

\begin{document}
\maketitle
\raggedbottom

\begin{abstract}
\noindent RGB-based video object segmentation (VOS) remains brittle under fast motion and adverse illumination. Event cameras preserve signal under blur and challenging exposure, but provide little evidence when motion slows. Interaction can restore object state for subsequent propagation, yet promptable event methods merely address single windows.
We formalize \textbf{e}vent-based \textbf{P}romptable \textbf{V}ideo \textbf{S}egmentation (ePVS), where prompts at arbitrary windows initialize, correct, or introduce objects and predictions. We present \textbf{E}vent \textbf{M}emory-\textbf{A}s-\textbf{P}rompt \textbf{S}egmentation (E-MAPS), which transfers a pretrained promptable segmenter to event streams and learns propagation and correction. Event-to-Token (E2T) aligns recurrent event features with pretrained representations while preserving prompt-conditioning priors. Memory-Aware Prompting (MAP) converts object-conditioned history into sparse prompts, allowing memory and user inputs to share one decoder pathway efficiently.
We construct \textbf{E}vent-\textbf{Vi}ew \textbf{M}asklets (EViM) from 206K aligned RGB--event pairs and $\sim$620K masks for two-stage representation transfer and native event-video learning, and establish three training-disjoint benchmarks. E-MAPS leads prompt-once and interactive ePVS across all three benchmarks while being more efficient. It further demonstrates strong performance on zero-shot object tracking benchmarks. Project page: \url{https://github.com/adonaydem/emaps-wacv}.
\end{abstract}

\section{Introduction} \label{sec:intro}
\noindent Video Object Segmentation (VOS) aims to segment target objects across video frames while maintaining their identities \cite{osvos,masktrack,stm,stcn,xmem,aot,cutie,rmem}.
Despite strong progress, RGB-based methods can struggle with fast motion and difficult lighting, where blur and poor exposure weaken visual cues \cite{vs,vos1,lim1,eventsam,event_sam_adapter,event_lowlight_vos,ecos,memorysam,evsam,seal}.
Event cameras address some of these challenges by recording brightness changes with microsecond latency and high dynamic range \cite{events_survey,events_survey_2}.
Yet their continuous, change-driven streams require efficient processing and may contain limited shape or identity cues during slow motion or low illumination \cite{events_survey_3,event_det}.
Tracking objects from events therefore requires maintaining identity through sparse activity, occlusion, and disappearance.
Promptable segmentation offers a flexible way to specify and correct objects using points, boxes, or masks \cite{sam1,tinysam,sami2v,sam2,sam3}.
For event streams, we find that these prompts can also recover object state and guide subsequent propagation when observations are weak.
Existing Event-SAM adaptations mostly focus on individual windows or RGB--event fusion \cite{eventsam,event_sam_adapter,seal,evsam}, while event-only VOS has mainly supported initial-mask propagation without subsequent interaction \cite{ecos}.
We therefore study an event-only setting that combines prompt-conditioned segmentation with temporal propagation in a single model.

We formalize \textbf{e}vent-based \textbf{P}romptable \textbf{V}ideo \textbf{S}egmentation (ePVS), a continuous event-stream task where point, box, or mask prompts may initialize, correct, or introduce objects at arbitrary windows, producing prompt-aligned masklets over time.
ePVS covers prompt-once propagation from first-window prompts and interactive segmentation, where later prompts correct existing objects.
Both settings require temporal propagation and the integration of later prompts into subsequent predictions.
We address this task with \textbf{E}vent \textbf{M}emory-\textbf{A}s-\textbf{P}rompt \textbf{S}egmentation (E-MAPS), an event-only framework combining TinySAM's compact promptable backbone with recurrent event encoding and a temporal memory \cite{tinysam}.
Its Event-to-Token (E2T) adapter maps recurrent event features onto the pretrained token grid.

We train E-MAPS progressively, first adapting pretrained representations
to event inputs and then learning to track objects over time.
In Stage I, aligned RGB observations guide E2T, while point, box,
and mask prompts preserve the pretrained model's segmentation behavior.
Stage II then learns mask propagation.
To maintain object identity across windows, Memory-Aware Prompting (MAP)
condenses past scene and object information into a fixed set of sparse
prompts for each target.
These enter the same decoder as user prompts, enabling initialization,
propagation, correction, and new-object insertion without repeatedly
processing full spatial memories.
To support both stages, we introduce \textbf{E}vent-\textbf{Vi}ew
\textbf{M}asklets (EViM), comprising 206{,}811 aligned RGB--event
pairs and approximately 620K masks from DSEC-Detection
\cite{dsec1,dsec2} and N-ImageNet \cite{nimagenet}, constructed
through cross-modal proposal transfer, geometric mapping,
and human quality control.

\noindent Our contributions are:
\begin{itemize}

    \item \textbf{Task.} We formalize ePVS, where prompt-once propagation and arbitrary-window interaction produce identity-consistent masklets despite intermittent event observations (Sec.~\ref{sec:task}).

    \item \textbf{Model.} We propose E-MAPS, an efficient promptable event-based VOS framework that uses E2T for transfer into a SAM-style feature space, learns propagation and interaction from event masklets, and injects temporal memory as sparse prompt tokens through MAP (Secs.~\ref{sec:adapter_prompt}--\ref{sec:training}).

    \item \textbf{Data.} We introduce EViM, with over 206K aligned RGB--event pairs and 620K masks supporting large-scale and diverse pretraining across both stages(Secs.~\ref{sec:data} and~\ref{sec:exp_setup}).

\end{itemize}

\noindent Experiments on EViM, FE108 \cite{fe108}, and VisEvent \cite{visevent} show that E-MAPS achieves leading prompt-once and interactive ePVS results. It outperforms SAM2.1 B+ while operating at a much faster frame rate. On the zero-shot VisEvent testbed, E-MAPS also achieves the strongest prompted result and the largest relative gain from additional prompts over SAM2 and SAM-I2V.

\section{Related Work}

\subsection{Promptable Segmentation Foundation Models}

\noindent
Promptable segmentation allows users to specify and refine target objects through points, boxes, or masks \cite{sam1}.
SAM established a flexible encoder--prompt--decoder architecture \cite{sam1}, which TinySAM later made more efficient \cite{tinysam}.
Building on this foundation, SAM2 introduced streaming memory for video segmentation \cite{sam2}, while SAM-I2V adapted image-based SAM models to promptable video tasks \cite{sami2v}.
Alongside these developments, VOS methods have relied on dense space--time memories to associate objects across frames and propagate their masks \cite{stm,stcn,xmem,aot,cutie,rmem}.
E-MAPS builds on TinySAM's compact semantic and prompt priors while learning temporal dynamics directly from event streams.

\vspace{-0.7em}
\subsection{Event Representation Transfer and SAM Adaptation}
\vspace{-0.25em}
\noindent
Event segmentation spans event-native encoders, RGB-to-event transfer, and cross-modal fusion \cite{sun2022ess,evsegnet,eisnet,cmx,cmnext}.
EventSAM enables SAM-like event prompting through weighted distillation \cite{eventsam}, while SAM-Event-Adapter and EVSAM inject event cues into image models \cite{event_sam_adapter,evsam}.
MemorySAM applies SAM2-style memory to multimodal semantic segmentation, and other work studies language guidance and event-domain representation learning \cite{memorysam,seal,openess,chen2026scaleevent,polizzi2026realm}.
These methods establish promptable event segmentation and cross-modal adaptation, mainly for per-window or multimodal prediction, while we seek to learn temporal dynamics under intermittent event streams.

\vspace{-0.5em}
\subsection{Event-based Video Segmentation and Tracking}
\vspace{-0.1em}
\noindent
ECOS is the closest event-only VOS predecessor, propagating an initial mask through recurrent event representations and cross-time matching \cite{ecos}.
Event-assisted low-light VOS~\cite{event_lowlight_vos} and EVOLVE~\cite{baek2025evolve} instead combine frames and events through memory matching or dual-memory refinement.
Related work also studies event-guided video segmentation, frame--event instance segmentation, and high-rate tracking through frame--event fusion or RGB-to-event transfer \cite{Yao_2025,Hamann_2025,zhang2023frameevent,wang2024eventvot}.
E-MAPS targets event-only mask propagation from point, box, or mask prompts, with arbitrary-window correction and object insertion.

\begin{figure}[t]
\centering \small
\includegraphics[width=0.45\textwidth]{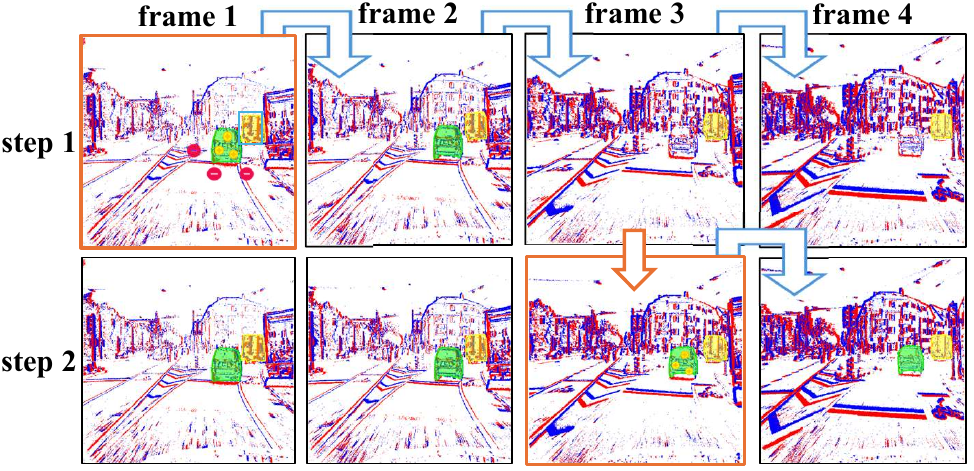}

{\small
\caption{\textbf{Interactive promptable segmentation with E-MAPS on an event sequence.}
\textbf{Step 1:} E-MAPS is prompted on frame~1 to segment two targets: a car
(\textcolor{dullgreen}{green}) with point prompts
(\textcolor{dullyellow}{orange}/\textcolor{dullred}{red} for positive/negative
clicks) and a truck (\textcolor{yellow}{yellow}) with a box
(\textcolor{dullcyan}{light blue}), then propagates masks to later windows
(\textcolor{dullblue}{blue} arrows). \textbf{Step 2:} when the car drifts after
event frame~2, a corrective prompt on frame~3 (\textcolor{dullred}{red} arrow)
recovers its masklet, while the truck continues without refinement.}
\label{fig:task_demo}
}

\vspace{-2em}
\end{figure}
\begin{figure*}[!t]
\centering \small
\includegraphics[width=0.9\textwidth,height=0.65\textheight,keepaspectratio]{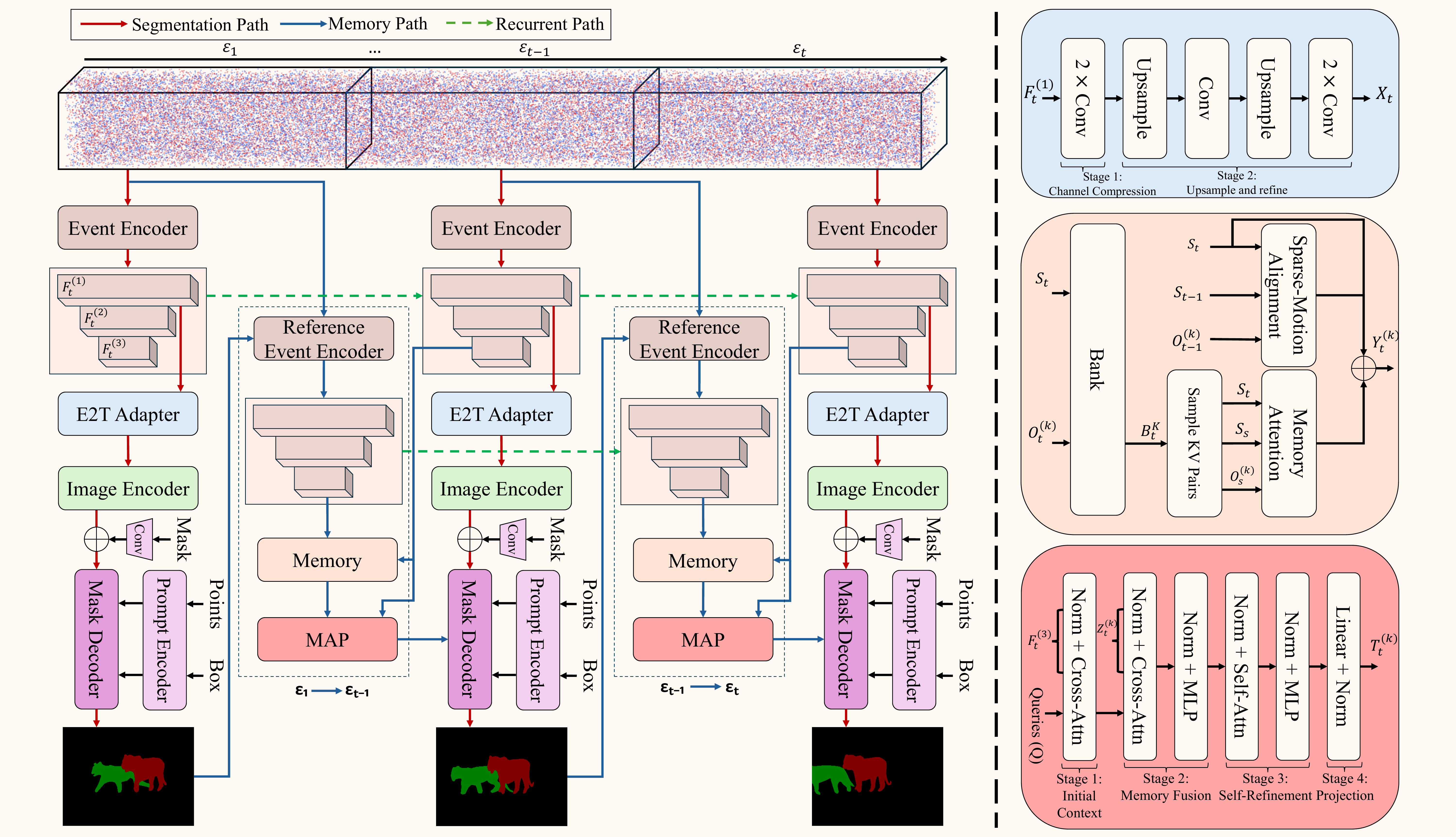}
{\small \caption{
E-MAPS overview. We encode event windows and adapt them into SAM-style tokens for promptable mask decoding. A temporal memory is converted into a small set of memory prompt tokens (MAP) and injected into the sparse-prompt pathway to support prompt-once propagation and interactive correction.
}
\label{fig:overview}
}
\vspace{-2em}
\end{figure*}

\section{Task: ePVS}
\label{sec:task}

\noindent \textbf{Event stream and temporal windows.}
An event camera produces an asynchronous stream 
$\mathcal{E}=\{e_i\}_{i=1}^{N}$, where each event 
$e_i=(x_i,y_i,t_i,p_i)$ records a brightness change at pixel location 
$(x_i,y_i)$, timestamp $t_i$, and polarity $p_i\in\{-1,+1\}$.
We standardize a stream as a group of ordered, non-intermittent and non-overlapping temporal windows:
\[
    \mathcal{E}_t =
    \{ e_i \in \mathcal{E} \mid \tau_t \leq t_i < \tau_{t+1} \},
    \qquad t=0,\ldots,T-1 .
\]
For a general task, a model can use any arbitrary event representation technique such as voxelization or event frames.

\noindent \textbf{Promptable event VOS.}
Given the event history and the prompts observed so far, the goal is to predict one mask for each active object identity at every window:
\[
    \{\hat{\mathbf{M}}_t^{(k)}\}_{k\in\mathcal{K}_t}
    =
    f_\theta(\mathcal{E}_{0:t}, \mathcal{P}_{0:t}).
\]
Here, $\mathcal{K}_t$ is the set of object identities currently being tracked, and $\mathcal{P}_{0:t}$ contains all user prompts issued up to time $t$. A prompt can initialize a new object, correct an existing object, or disambiguate the target when multiple objects are present. Fig.~\ref{fig:task_demo} illustrates this protocol where objects are initialized with prompts, propagated over time, and selectively corrected when drift or missed detections occur.

\section{Methodology} \label{sec:method}

\subsection{Event Window Representation and Encoding} \label{sec:representation}

\noindent \textbf{Event Representation.}
We partition the stream $\mathcal{E}$ into temporal windows $\{\mathcal{E}_t\}_{t=0}^{T-1}$ of duration $\Delta T$ and convert each window into a signed voxel volume $\mathbf{V}_t\in\mathbb{R}^{B\times H\times W}$~\cite{Zhu_2019_CVPR}.
Events are assigned to $B$ temporal bins using the normalized timestamp
\[
\tau_i=\frac{B-1}{\Delta T}(t_i-t_t^{\mathrm{start}}),
\]
and accumulated with linear temporal interpolation:
\[
\mathbf{V}_t[b,y_i,x_i]
\mathrel{+}=
p_i\,\max\!\left(0,1-\left|b-\tau_i\right|\right).
\]
Here, \(b\) indexes a temporal bin of the voxel grid, \(p_i\in\{-1,+1\}\) is the event polarity, and \((x_i,y_i)\) denotes the integer pixel location of the event. Unless stated otherwise, we use $\Delta T{=}50$ms and $B{=}5$, and apply no voxel normalization since event-density variation is handled by our encoder.

\noindent \textbf{Recurrent Temporal Embedding Extraction (RTEE).}
RTEE is our temporal feature extractor inspired by ECOSNet~\cite{ecos}, a recurrent ConvLSTM encoder (ResLSTM) that maps voxels to pyramid features. We use two RTEE blocks: an \emph{event encoder} for the current window and a \emph{reference event encoder} that converts each window and its predicted mask into an object-conditioned memory feature:
\begin{align}
\mathbf{F}^{\mathrm{evt}}_t &= \mathrm{RTEE}_{\mathrm{evt}}\!\left(\mathbf{V}_t,\ \mathrm{state}^{\mathrm{evt}}_{t-1}\right), \\
\mathbf{F}^{\mathrm{ref},(k)}_{t} &= \mathrm{RTEE}_{\mathrm{ref}}\!\left(\mathbf{V}_{t},\ \hat{\mathbf{M}}_{t}^{(k)},\ \mathrm{state}^{\mathrm{ref}}_{t-1}\right),
\end{align}
where $\mathbf{V}_t$ denotes the event voxel, $\mathrm{state}^{(\cdot)}_{t-1}$ the hidden states across recurrent temporal stages, and $\hat{\mathbf{M}}_{t}^{(k)}$ the current mask prediction. $\mathbf{F}^{\mathrm{evt}}_t$ is a multi-scale pyramid containing $\mathbf{F}^{(1)}_t,\mathbf{F}^{(2)}_t,\mathbf{F}^{(3)}_t$. The reference feature is stored for subsequent windows.

\subsection{Token Alignment and Promptable Decoding} \label{sec:adapter_prompt}

\noindent \textbf{Event-to-Token (E2T) adapter.}
To reuse a pretrained backbone's rich semantic representation without retraining the full token encoder and decoder from scratch, we introduce an E2T adapter that maps the shallow event feature $\mathbf{F}^{(1)}_t$ to the image encoder feature space and replaces the SAM patch embedding. E2T, shown as the blue block in Fig.~\ref{fig:overview}, maps this feature to the token-grid input used by the promptable backbone
\begin{equation}
\mathbf{X}_t
=
{\mathrm{E2T}}\!\left(\mathbf{F}^{(1)}_t\right).
\end{equation}
Here, $\mathbf{X}_t$ serves as the patch embedding. E2T first compresses the feature channels. It then applies two upsampling stages with convolutional refinement to match the spatial resolution and channel width of the pretrained token grid.

\noindent \textbf{Image encoder, prompt encoder, and mask decoder.}
We use TinySAM~\cite{tinysam} for its efficient pretrained point-, box-, and mask-prompt interface. Temporal association is learned from scratch through the event memory described below. The image encoder processes $\mathbf{X}_t$ directly to produce shared feature embeddings.
For object $k$, the prompt encoder maps the prompt set $\mathcal{P}_t^{(k)}$ into sparse and dense prompt embeddings.
The TinySAM mask decoder combines the shared feature embeddings and the object prompt embeddings to predict a mask.
To handle object absence and heavy occlusion, we add an occlusion token and a lightweight occlusion head following SAM2~\cite{sam2}. The head is trained with focal loss using empty masks as absent-object labels.

\subsection{Temporal Memory as Prompts} \label{sec:memory_map}

\noindent \textbf{Memory bank.}
For temporal propagation, we maintain a memory bank of past scene and object features. Let $\mathbf{S}_t=\mathbf{F}^{(3)}_t\in\mathbb{R}^{C\times H_m\times W_m}$ denote the deepest scene feature extracted by the event encoder at window $t$. For object $k$, let $\mathbf{O}^{(k)}_t\in\mathbb{R}^{C\times H_m\times W_m}$ denote the deepest feature produced by the reference event encoder in Sec.~\ref{sec:representation}. It carries object-specific information from the preceding event window and predicted mask. Let $\mathcal{I}_t$ contain the retained memory indices. These include the first prompted window and the $s-1$ most recent windows. At time $t$, the memory bank for object $k$ is
\begin{equation}
\mathcal{B}^{(k)}_t
=
\{(\mathbf{S}_j,\mathbf{O}^{(k)}_j)\}_{j\in\mathcal{I}_t},
\qquad
|\mathcal{I}_t|\leq s.
\end{equation}
Each entry pairs scene context with the object-conditioned feature stored at the same memory index.

\noindent \textbf{Memory attention.}
Given the current scene feature $\mathbf{S}_t$, we retrieve object-specific and global temporal evidence from $\mathcal{B}^{(k)}_t$ using standard attention~\cite{vaswani2023attentionneed}. The sequences from the retained feature entries are stacked along the memory dimension. The current scene feature provides the query. The stored scene features provide the keys, and the stored object features provide the values
\begin{equation}
\begin{aligned}
\mathbf{q}_t &= f_q(\mathbf{S}_t),\\
\mathbf{K}_t &= \{f_k(\mathbf{S}_j)\}_{j\in\mathcal{I}_t},\\
\mathbf{V}^{(k)}_t &= \{f_v(\mathbf{O}^{(k)}_j)\}_{j\in\mathcal{I}_t},
\end{aligned}
\end{equation}
where $f_q$, $f_k$, and $f_v$ are learned projections. The memory readout is
\begin{equation}
\mathbf{R}^{(k)}_t
=
\mathrm{softmax}
\left(
\frac{\mathbf{q}_t\mathbf{K}^{\top}_t}{\sqrt{d}}
\right)
\mathbf{V}^{(k)}_t.
\end{equation}

\noindent \textbf{Sparse-Motion Adjustment.}
To improve robustness under extreme sparsity and abrupt motion, we align the most recent stored object feature with the current window using an offset-conditioned deformable convolution following ECOS~\cite{ecos,dai2017deformableconvolutionalnetworks}:
\begin{equation}
\mathbf{R}^{(k)}_{t,\mathrm{sparse}}
=
\mathrm{DeformConv}\!\left(
\mathbf{O}^{(k)}_{t-1},
f_{\delta}\!\left([\mathbf{S}_t,\mathbf{S}_{t-1}]\right)
\right),
\end{equation}
where $[\cdot,\cdot]$ denotes channel-wise concatenation and $f_{\delta}$ predicts the offsets. This produces an aligned object feature that better matches the current window under sparse observations or abrupt motion. The final memory output is
\begin{equation}
\mathbf{Y}^{(k)}_t
=
\mathbf{S}_t
+
\mathbf{R}^{(k)}_t
+
\mathbf{R}^{(k)}_{t,\mathrm{sparse}}.
\end{equation}

If the memory bank is empty, we set $\mathbf{Y}^{(k)}_t=\mathbf{S}_t$. MAP converts $\mathbf{Y}^{(k)}_t$ into sparse memory prompt tokens for the mask decoder.

\noindent \textbf{Memory-Aware Prompting (MAP).}
A key design choice is how to use retrieved object history to guide segmentation.
We explore three approaches: past-object feature fusion, dense memory cross-attention~\cite{sam2}, and prompt-based memory injection.
Past-object fusion uses only the latest stored object feature, removing temporal attention.
Dense cross-attention instead incorporates the event-derived memory readout through a randomly initialized SAM2-style decoder, without relying on RGB features or pretrained SAM2 memory weights.
As shown in Sec.~\ref{sec:ablations}, prompt-based injection offers the best balance between segmentation accuracy and computational cost.
This design also fits naturally with the promptable decoder.
While the retrieved memory $\mathbf{Y}^{(k)}_t$ contains dense object-conditioned information, the decoder receives object guidance through sparse prompts.
MAP bridges these representations by summarizing the retrieved history into $K$ memory prompt tokens $\mathbf{T}^{(k)}_t\in\mathbb{R}^{K\times C}$, where $C$ denotes the token width.
Following compact conditioning approaches for spatial and spatio-temporal segmentation~\cite{glamm,videoglamm,videolisa}, MAP uses $K$ learnable queries $\mathbf{P}_{\mathrm{map}}\in\mathbb{R}^{K\times C}$ and spatial tokens $\mathbf{Z}^{(k)}_t$ formed by channel-wise fusion of $\mathbf{Y}^{(k)}_t$ and $\mathbf{F}^{(3)}_t$.
MAP computes
\begin{align}
\widehat{\mathbf{P}}_{\mathrm{map}}
&=
\mathbf{P}_{\mathrm{map}}
+
\mathrm{Attn}\!\left(
\mathbf{P}_{\mathrm{map}},
\mathbf{F}^{(3)}_t,
\mathbf{F}^{(3)}_t
\right),\\
\widetilde{\mathbf{T}}^{(k)}_t
&=
\widehat{\mathbf{P}}_{\mathrm{map}}
+
\mathrm{Attn}\!\left(
\widehat{\mathbf{P}}_{\mathrm{map}},
\mathbf{Z}^{(k)}_t,
\mathbf{Z}^{(k)}_t
\right),\\
\mathbf{T}^{(k)}_t
&=
\widetilde{\mathbf{T}}^{(k)}_t
+
\mathrm{FFN}\!\left(
\widetilde{\mathbf{T}}^{(k)}_t
\right).
\end{align}
Here, $\mathrm{Attn}(\mathbf{q},\mathbf{k},\mathbf{v})$ denotes multi-head attention with learned query, key, and value projections. The attention and FFN blocks use pre-normalization and residual connections. For $K>1$, self-attention and an FFN further refine $\mathbf{T}^{(k)}_t$. We use $K=2$ by default.

At a prompted window, $\mathbf{T}^{(k)}_t$ is concatenated with the sparse user-prompt embeddings before mask decoding. At an unprompted window, $\mathbf{T}^{(k)}_t$ is the only input to the sparse-prompt pathway.

\subsection{Training} \label{sec:training}
\noindent We train E-MAPS in two stages: (I) token-space alignment and (II) video training with memory prompting.
\noindent \textbf{Stage I: Token-space alignment.}
Stage I aligns event tokens to a pretrained image encoder using only single frame/window images.

\vspace{-1em}
\noindent \paragraph{Sub-stage I-a (embedding-space bootstrapping).}
We freeze the pretrained SAM modules and train only RTEE+E2T.
Let \(\Phi\) be the pretrained image encoder excluding patch embedding, and let \(\Phi^{(\ell)}\) denote its output after block \(\ell\). Given E2T event tokens \(\mathbf{X}_t\) and aligned RGB teacher tokens \(\mathbf{Z}_t\), we compute
\begin{equation}
\hat{\mathbf{E}}^{(\ell)}_t=\Phi^{(\ell)}(\mathbf{X}_t), \qquad
\mathbf{E}^{(\ell)}_t=\Phi^{(\ell)}(\mathbf{Z}_t),
\end{equation}
where \(\hat{\mathbf{E}}^{(\ell)}_t\) and \(\mathbf{E}^{(\ell)}_t\) are event and RGB teacher features, respectively.

We apply \(L_1\) feature distillation at an intermediate block \((\ell{=}2)\) and the final block \((\ell{=}3)\):
\begin{equation}
\mathcal{L}_{\mathrm{emb}}
=
\|\hat{\mathbf{E}}^{(2)}_t-\mathbf{E}^{(2)}_t\|_1
+
\lambda\|\hat{\mathbf{E}}^{(3)}_t-\mathbf{E}^{(3)}_t\|_1 .
\end{equation}

We also use InfoNCE~\cite{cpc, radford2021learningtransferablevisualmodels} between final-block event and RGB teacher features:
\begin{equation}
\mathcal{L}_{\mathrm{NCE}}
=
\mathrm{InfoNCE}\!\left(\hat{\mathbf{E}}^{(3)}_t,\mathbf{E}^{(3)}_t\right),
\end{equation}
We further use CORAL, inspired by Deep CORAL~\cite{deepcoral}, inorder to align second-order statistics between E2T features and RGB teacher patch embeddings $\mathbf{Z}_t$:
\begin{equation}
\mathcal{L}_{\mathrm{cov}}
=
\mathrm{CORAL}\!\left(\mathbf{X}_t,\mathbf{Z}_t\right).
\end{equation}

To preserve TinySAM prompt behavior, we decode masks from sampled prompts and apply \(\mathcal{L}_{\mathrm{Dice}}\). The Sub-stage I-a objective is:
\begin{equation}
\mathcal{L}_{\mathrm{DA}}
=
\mathcal{L}_{\mathrm{emb}}
+
\gamma\mathcal{L}_{\mathrm{cov}}
+
\delta\mathcal{L}_{\mathrm{NCE}}
+
\eta\mathcal{L}_{\mathrm{Dice}},
\end{equation}
where $\gamma$, $\delta$, $\eta$ denote weights for the corresponding objectives.
\vspace{-1em}
\noindent \paragraph{Sub-stage I-b (full adaptation).}
Following similar configuration of Sub-stage I-a, we unfreeze the encoder and mask decoder and continue training.

\noindent \textbf{Stage II: Video training with memory prompting.}
We train on clips \((T>1)\) with our MAP-based memory (Sec.~\ref{sec:memory_map}), with all parameters unfrozen. We supervise masks with Dice and focal losses, occlusion with focal BCE using \(o_t{=}1\) when the ground-truth mask is empty, and mask confidence with an \(L_1\) IoU loss following TinySAM:
\begin{equation}
\mathcal{L}_{\mathrm{video}}
=
\mathcal{L}_{\mathrm{Dice}}
+
\nu\mathcal{L}_{\mathrm{focal}}
+
\xi\mathcal{L}_{\mathrm{occ}}
+
\beta\mathcal{L}_{\mathrm{iou}},
\end{equation}
where $\nu$, $\xi$, $\beta$ denote weights for the corresponding objectives.
Here, \(\mathcal{L}_{\mathrm{focal}}\) is the mask focal loss, \(\mathcal{L}_{\mathrm{occ}}\) is the occlusion focal BCE loss, and \(\mathcal{L}_{\mathrm{iou}}\) regresses predicted mask IoU to learn confidence.

\noindent \paragraph{Interactive training and scheduled teacher forcing.}
During Stage II, we simulate interactive corrections by injecting foreground and
background clicks whenever
$\operatorname{Dice}(\hat{\mathbf{M}}_t,\mathbf{M}_t)$ falls below a threshold.
Since early noisy predictions can contaminate memory and amplify errors, we use
scheduled teacher forcing, where memory writes initially use ground-truth masks, then
gradually transition to predicted masks $\hat{\mathbf{M}}^{(k)}_t$ under a
decaying schedule.

\begin{table}[!t]
\centering
\caption{Comparison of event-camera datasets and EViM.}
\label{tab:data_stats}
\vspace{-0.5em}
\setlength{\tabcolsep}{3pt}
\renewcommand{\arraystretch}{0.95}

\resizebox{\columnwidth}{!}{%
\begin{tabular}{lcccl}
\toprule
\rowcolor{Gray}
Dataset / Source & Frames & Masks & Task & Image/Video \\
\midrule
DDD17-Seg~\cite{evsegnet}
              & 19{,}840 & -- & Semantic & Image \\
DSEC-Semantic~\cite{sun2022ess}
              & 10{,}891 & -- & Semantic & Image \\
M3ED-Semantic~\cite{li2025efficienteventbasedsemanticsegmentation}
              & 7{,}275 & -- & Semantic & Image \\
RGBE-SEG~\cite{eventsam}
              & 65{,}957 & -- & Instance & Image \\
MouseSIS~\cite{Hamann_2025}
              & -- & 75{,}532 & Instance & Video \\
\midrule
\rowcolor{gaincolor}
\multicolumn{5}{l}{\textbf{EViM (ours)}} \\
DSEC-Detection~\cite{dsec1,dsec2}
              & 70{,}379 & 338{,}000 & Instance & Video \\
N-ImageNet~\cite{nimagenet}
              & 136{,}432 & 281{,}869 & Instance & Image \\
\rowcolor{gaincolor}
\textbf{Total}
              & \textbf{206{,}811}
              & \textbf{$\sim$620K}
              & \textbf{Instance}
              & \textbf{Image \& Video} \\
\bottomrule
\end{tabular}%
}
\vspace{-0.5em}
\end{table}

\section{Data Construction} \label{sec:data}

\subsection{Datasets and EViM construction.}
We build supervision from three complementary event sources. EOS~\cite{ecos} provides DAVIS-style sequences with native event-view instance masks. DSEC-Detection~\cite{dsec1,dsec2} provides real driving videos with dense boxes, class labels, and persistent track IDs. N-ImageNet~\cite{nimagenet} contributes broad semantic diversity through RGB--event pairs. Since DSEC-Detection and N-ImageNet lack instance masks, we construct and release \textbf{E}vent-\textbf{Vi}ew \textbf{M}asklets (EViM) using a semi-automated cross-modal annotation pipeline. EViM contains 206{,}811 RGB--event pairs and approximately 620K event-view instance masks. Table~\ref{tab:data_stats} compares EViM with prior datasets, which mainly provide semantic labels, image-level instances, or smaller video collections.

\noindent \textbf{EViM construction.}
For each RGB--event pair, an RGB teacher model generates candidate masks with confidence scores. Accepted masks are then transferred into the event-camera image plane. For teacher model, we consider large variants of SAM2.1~\cite{sam2} and SAM3~\cite{sam3}. SAM3 produces higher-quality masks, while cross-teacher agreement shows limited dependence on the selected teacher.

For DSEC-Detection, we prompt SAM3 on the synchronized RGB image using the ground-truth box and class text. We project each mask to the left event-camera view using the provided calibration. Masks are retained based on confidence and box--mask agreement, then grouped by track ID to form masklets.

For N-ImageNet~\cite{nimagenet}, we prompt SAM3 using ImageNet~\cite{imagenet}
synset labels and co-occurring concepts to improve multi-object coverage.
Since the RGB and event views are not aligned, we estimate an
RGB$\rightarrow$event homography.
RoMa-MINIMA~\cite{roma,minima} provides tentative correspondences
$\mathcal{C}=\{(\mathbf{x}_i,\mathbf{u}_i)\}_{i=1}^{N}$,
where $\mathbf{x}_i$ and $\mathbf{u}_i$ are matched points in the RGB
and event images, respectively, and $N$ is the number of matches.
We estimate the homography $\hat{H}=\operatorname{RANSAC}(\mathcal{C})$
and evaluate its geometric consistency using the inlier ratio
\begin{equation}
\rho =
\frac{
\left|
\left\{i \,\middle|\,
\|\mathbf{u}_i-\pi(\hat{H}\tilde{\mathbf{x}}_i)\|_2
<\tau_{\mathrm{geo}}
\right\}
\right|
}{N},
\end{equation}
where $\tilde{\mathbf{x}}_i$ denotes homogeneous RGB coordinates,
$\pi(\cdot)$ maps them back to 2D coordinates, and
$\tau_{\mathrm{geo}}=5$ pixels is the reprojection-error threshold.
The ratio $\rho$ measures the fraction of correspondences consistent
with the estimated alignment. Masks whose teacher score and $\rho$ exceed fixed thresholds are warped into the event view by inverse mapping
\begin{equation}
M^E(\mathbf{u})
=
M^I\!\left(
\pi\!\left(\hat{H}^{-1}\tilde{\mathbf{u}}\right)
\right).
\end{equation}
Here, $M^I$ and $M^E$ denote the RGB and event-view masks. This planar model is suitable because N-ImageNet displays source images on a monitor while the event camera moves approximately parallel to the screen.

\noindent\textbf{EViM benchmarks.} We construct three fixed, training-disjoint benchmarks from the DSEC-Detection test split. EViM500 and EViM1000 respectively contain approximately 4.7K/3K clips and 47K/52K object instances, spanning 350--500\,ms and up to 1\,s. Their masks pass automatic QC and stricter human verification. EViM-HM provides fully verified long-horizon evaluation. 

\subsection{Quality control (QC)}
EViM combines automatic filtering with human verification. DSEC proposals are filtered using confidence and box--mask agreement, while N-ImageNet proposals are filtered using confidence and homography support. Reviewers then audit RGB--event--mask triplets for object identity, event-view alignment and geometric validity. In a final 2{,}000-triplet audit, 92.2\% of post-QC masks are judged correct, while the hard failure rate decreases from 25.7\% before QC to 4.8\%. 

\begin{table}[!b]

\centering

\caption{Prompt-once VOS on EViM500, EViM1000, and EViM-HM, reported using $\mathcal{J}\&\mathcal{F}$. Parentheses denote gains over the second-best result. \emph{ft} indicates models fine-tuned on our dataset.}

\label{tab:vos_evim}

\vspace{-0.5em}

\setlength{\tabcolsep}{3pt}

\renewcommand{\arraystretch}{0.95}

\small

\resizebox{\columnwidth}{!}{%

\begin{tabular}{lllcccccc}

\toprule

\rowcolor{Gray}

Method & Rep. & Params (M) $\downarrow$ & EViM500 & EViM1000 & EViM-HM
& FPS $\uparrow$ & GFLOPs/frame $\downarrow$ \\

\midrule

SAM-I2V~\cite{sami2v}
& Recon.
& \textcolor{red}{18.90}
& 79.90
& 73.81
& 69.92
& 9.00
& 211.43 \\

SAM2.1 B+~\cite{sam2}
& Recon.
& 80.85
& 80.01
& 74.45
& 70.48
& 2.65
& 390.00 \\

XMem++~\cite{bekuzarov2023xmem}
& Recon.
& 63.30
& 80.84
& 72.46
& 68.55
& 8.40
& 229.57 \\

SwinB-AOT-L~\cite{aot}
& Recon.
& 65.40
& 77.96
& 70.32
& 66.41
& 14.38
& \textbf{\textcolor{red}{82.29}} \\

SwinB-DeAOT-L~\cite{deaot}
& Recon.
& 70.30
& \textbf{\textcolor{blue}{82.84}}
& 76.38
& 72.44
& \textcolor{blue}{18.45}
& \textbf{\textcolor{green}{87.00}} \\

EventSAM+SAM2.1 B+~\cite{eventsam,sam2}
& Event voxels+Recon.
& 174.55
& 82.16
& 76.91
& 72.96
& 2.60
& 602.57 \\

EventSAM+SwinB-DeAOT-L~\cite{eventsam,deaot}
& Event voxels+Recon.
& 164.00
& \textbf{\textcolor{green}{83.17}}
& \textbf{\textcolor{green}{77.42}}
& \textbf{\textcolor{blue}{73.49}}
& 8.01
& \textbf{\textcolor{blue}{100.57}} \\

ECOS~\cite{ecos}
& Event voxels
& \textcolor{green}{31.47}
& 69.22
& 60.13
& 56.08
& \textbf{\textcolor{red}{38.15}}
& 106.43 \\

\midrule

ECOS ft~\cite{ecos}
& Event voxels
& \textcolor{green}{31.47}
& 79.09
& 76.67
& \textbf{\textcolor{green}{75.71}}
& \textbf{\textcolor{red}{38.15}}
& 106.43 \\

SAM-I2V ft
& Recon.
& \textcolor{red}{18.90}
& 81.96
& 75.31
& 72.35
& 9.00
& 211.43 \\

SAM2.1 B+ ft
& Recon.
& 80.85
& 82.82
& \textbf{\textcolor{blue}{77.06}}
& 72.08
& 2.65
& 390.00 \\

\midrule

\rowcolor{gaincolor}

E-MAPS (Ours)
& Event voxels
& \textcolor{blue}{36.25}
& \textbf{\textcolor{red}{84.38}}
  {\tiny(\textcolor{gray}{\textbf{1.2$\uparrow$}})}
& \textbf{\textcolor{red}{80.37}}
  {\tiny(\textcolor{gray}{\textbf{3.0$\uparrow$}})}
& \textbf{\textcolor{red}{76.41}}
  {\tiny(\textcolor{gray}{\textbf{0.7$\uparrow$}})}
& \textbf{\textcolor{green}{27.41}}
& 150.43 \\

\bottomrule

\end{tabular}%

}

\vspace{-0.5em}

\end{table}

\begin{figure*}[!t]
    \centering
    \includegraphics[width=\textwidth]{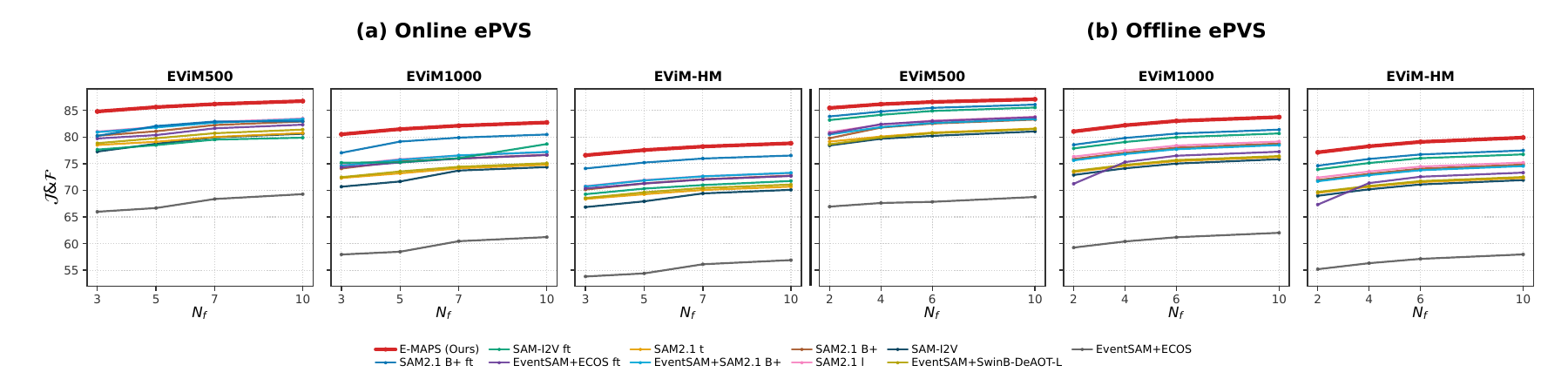}
    \caption{Interactive ePVS performance on EViM500, EViM1000, and EViM-HM under online and offline evaluation. Results are reported using $\mathcal{J}\&\mathcal{F}$ as the number of interaction frames $N_f$ increases.}
    \label{fig:epvs}
    \vspace{-0.5em}
\end{figure*}

\section{Experiments}
\label{sec:exp_setup}

\noindent \textbf{Training setup.}
We initialize the TinySAM image encoder, prompt encoder, and mask decoder from the public TinySAM checkpoint~\cite{tinysam}, whereas RTEE, E2T, the memory module, MAP, and the occlusion head are randomly initialized and trained from scratch. Training follows the two-stage procedure in Sec.~\ref{sec:training}: Stage I performs token-space alignment for 20 epochs with batch size 16, and Stage II trains clip-level propagation with memory prompting for 20 epochs using clips of 10 windows and memory window ($s$) is 3. We take 384x384 event voxels and optimize with AdamW on two A100 80GB GPUs, using cosine learning-rate decay with a 5\% linear warmup, weight decay 0.01, and betas 0.9, 0.999. Newly introduced modules use an initial learning rate of $3{\times}10^{-4}$, while TinySAM-initialized modules start from $1{\times}10^{-5}$. 

\noindent\textbf{Evaluation protocols.}
We evaluate four complementary settings on EViM and external benchmarks. EViM500, EViM1000, and EViM-HM are fixed, training-disjoint benchmarks using non-overlapping 50\,ms event windows and reporting $\mathcal{J}$, $\mathcal{F}$, and $\mathcal{J}\&\mathcal{F}$ (Sec.~\ref{sec:data}). Quantitative tables mark the top three entries per column in \textcolor{red}{red}, \textcolor{green}{green}, and \textcolor{blue}{blue}. In \emph{prompt-once VOS}, a box initializes each object in its first annotated window for propagation without further interaction. In \emph{interactive ePVS}, each corrected window receives three clicks. Online evaluation corrects predictions below 0.75 predicted IoU in a single pass, whereas offline evaluation first processes the sequence and then corrects the lowest-IoU windows under the same budget. FE108~\cite{fe108} and VisEvent evaluate zero-shot event-based box tracking using success rate (SR). Zero-shot interactive tracking on VisEvent allows up to $3$ corrections. Released checkpoints are used unless marked \emph{ft}, indicating full-parameter finetuning on the same EViM split and event-window schedule. \emph{Recon.} denotes E2VID~\cite{e2vid} reconstructions. 

\begin{table}[!t]
\centering
\caption{
Zero-shot VOT evaluation on FE108 and VisEvent using Success Rate (SR). EV denotes Event representation compatible with the corresponding method.
}
\label{tab:zero_shot}
\vspace{-0.5em}

\setlength{\tabcolsep}{3.2pt}
\renewcommand{\arraystretch}{1.02}
\scriptsize

\resizebox{\columnwidth}{!}{%
\begin{tabular}{@{}lc|c|cccc@{}}
\toprule
\rowcolor{Gray}
Method & Mod.
&
& \multicolumn{4}{c}{VisEvent} \\
\rowcolor{Gray}
&
& FE108
& LI & FOC+POC & FM+MB & BOM+BC \\
\midrule

SAM+XMem++~\cite{sam1,bekuzarov2023xmem}
& Frame
& 60.87
& 31.8
& 33.9
& 37.6
& 34.5 \\

SAM+SwinB-DeAOT-L~\cite{sam1,deaot}
& Frame
& 62.61
& \textcolor{blue}{33.7}
& 35.1
& 39.5
& 35.6 \\

SAM+SwinB-AOT-L~\cite{sam1,aot}
& Frame
& 62.41
& 32.9
& 34.4
& 38.8
& 35.1 \\

SAM-I2V~\cite{sami2v}
& Frame
& 61.23
& 30.7
& 32.7
& 36.2
& 33.2 \\

SAM2.1 B+~\cite{sam2}
& Frame
& \textcolor{red}{72.88}
& 32.6
& 30.3
& 31.6
& 30.6 \\

EventSAM+ECOS~\cite{eventsam,ecos}
& EV
& 65.25
& 27.4
& 28.6
& 34.3
& 28.9 \\

\midrule

\rowcolor{gaincolor}
E-MAPS (Ours)
& EV
& 70.15
& \textcolor{red}{36.1}
& \textcolor{green}{36.5}
& \textcolor{green}{43.2}
& \textcolor{green}{37.2} \\

\midrule

LLE-VOS
& Frame+EV
& 69.74
& 32.6
& \textcolor{blue}{36.1}
& 42.1
& 35.9 \\

EVOLVE
& Frame+EV
& \textcolor{green}{71.46}
& 30.9
& 35.8
& \textcolor{red}{43.9}
& \textcolor{blue}{36.6} \\

\rowcolor{gaincolor}
E-MAPS
& Frame+EV
& \textcolor{blue}{71.32}
& \textcolor{green}{35.7}
& \textcolor{red}{37.2}
& \textcolor{blue}{42.7}
& \textcolor{red}{37.7} \\

\bottomrule
\end{tabular}%
}

\vspace{-0.2em}
\end{table}

\vspace{\baselineskip}
\noindent\textbf{Prompt-once VOS and efficiency.}
E-MAPS consistently achieves the highest $\mathcal{J}\&\mathcal{F}$ across EViM500, EViM1000, and EViM-HM, with scores ranging from 76.41 to 84.38.
Even after fine-tuning competing methods such as ECOS, SAM-I2V, and SAM2.1 B+, E-MAPS retains its advantage across all three benchmarks.
Beyond segmentation accuracy, E-MAPS offers substantial computational benefits.
E-MAPS runs at 27.41 FPS, over ten times faster than SAM2.1 B+, while requiring considerably fewer operations per frame.
Together, these results demonstrate that E-MAPS achieves accurate and efficient event-based tracking, with qualitative examples in Fig.~\ref{fig:samples} further illustrating its robustness under low-light conditions.

\noindent\textbf{Interactive ePVS.}
E-MAPS consistently outperforms competing methods across all correction budgets in both online and offline protocols.
Performance improves steadily as additional corrections are introduced, with online $\mathcal{J}\&\mathcal{F}$ gains of approximately 2 points across all three benchmarks.
Offline interaction yields further improvements, reaching 87.08 on EViM500, 83.74 on EViM1000, and 79.89 on EViM-HM.
These results demonstrate that E-MAPS effectively incorporates new user prompts into its memory, allowing corrections to improve not only the current segmentation but also subsequent propagation.

\begin{table}[t]
\centering
\caption{Zero-shot interactive tracking on VisEvent. $\Delta_{\mathrm{rel}}$ denotes relative gain after corrections. NMO denotes no motion}
\label{tab:visevent_interactive}
\vspace{-0.5em}
\scriptsize
\setlength{\tabcolsep}{2.1pt}
\renewcommand{\arraystretch}{0.95}
\begin{tabular}{@{}lcccc@{}}
\toprule
\rowcolor{Gray}
Method & \multicolumn{2}{c}{Overall} & \multicolumn{2}{c}{NMO} \\
\rowcolor{Gray}
& $\mathrm{SR}_{0/3}$ & $\Delta_{\mathrm{rel}}$ & $\mathrm{SR}_{0/3}$ & $\Delta_{\mathrm{rel}}$ \\
\midrule
SAM-I2V ft~\cite{sami2v}
& 33.2/36.0 & 8.4\% & 26.7/29.4 & 10.1\% \\
SAM2.1 B+ ft~\cite{sam2}
& 31.3/34.4 & 10.0\% & 24.8/27.6 & 11.3\% \\
\rowcolor{gaincolor}
E-MAPS (Ours)
& \textbf{38.3/44.1} & \textbf{15.3\%} & \textbf{28.9/37.8} & \textbf{30.8\%} \\
\bottomrule
\end{tabular}
\vspace{-0.7em}
\end{table}

\begin{figure*}[!t]
\centering
\newcommand{\imgw}{0.13\textwidth}
\newcommand{\imgh}{1.55cm}

\setlength{\tabcolsep}{0pt}
\renewcommand{\arraystretch}{0.85}

\begin{tabular}{@{}cccccc@{}}
\includegraphics[width=\imgw,height=\imgh]{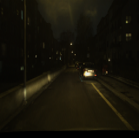} &
\includegraphics[width=\imgw,height=\imgh]{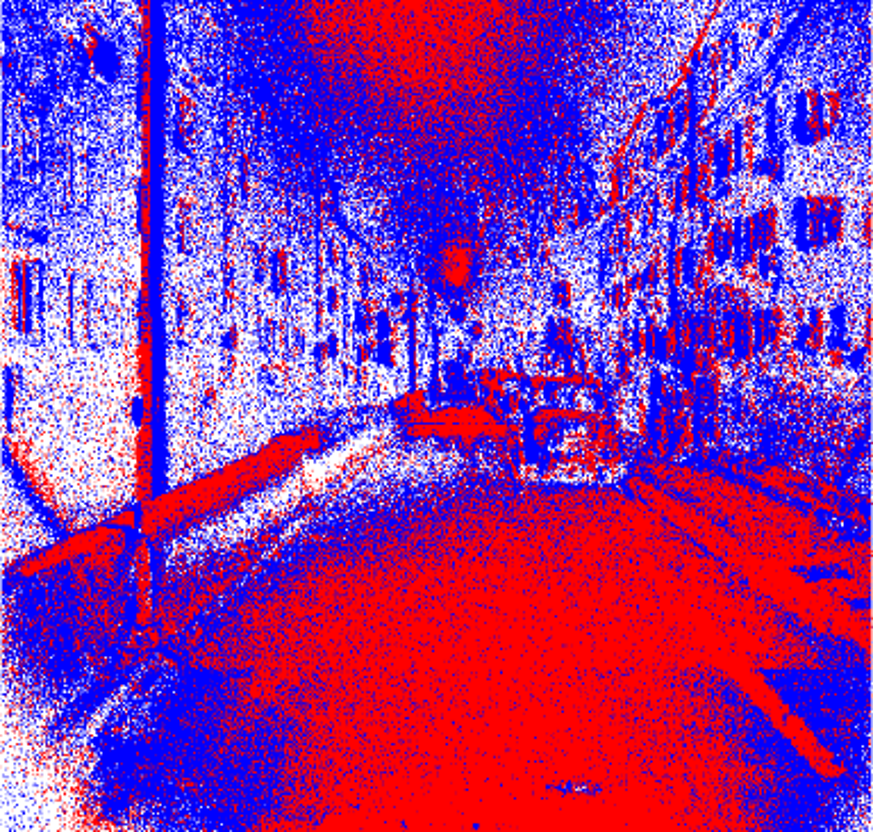} &
\includegraphics[width=\imgw,height=\imgh]{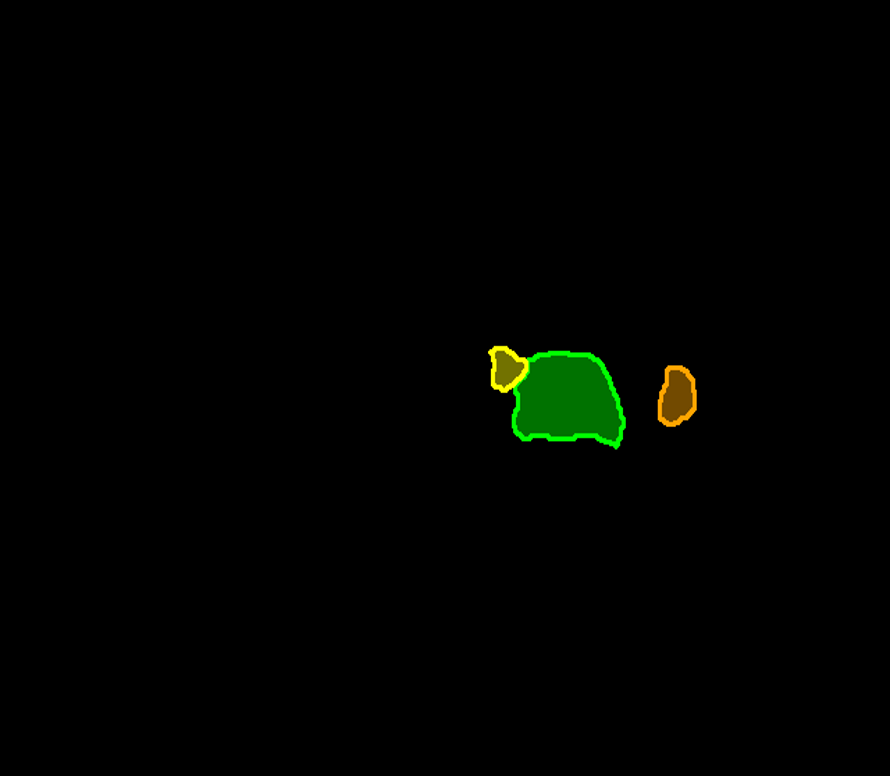} &
\includegraphics[width=\imgw,height=\imgh]{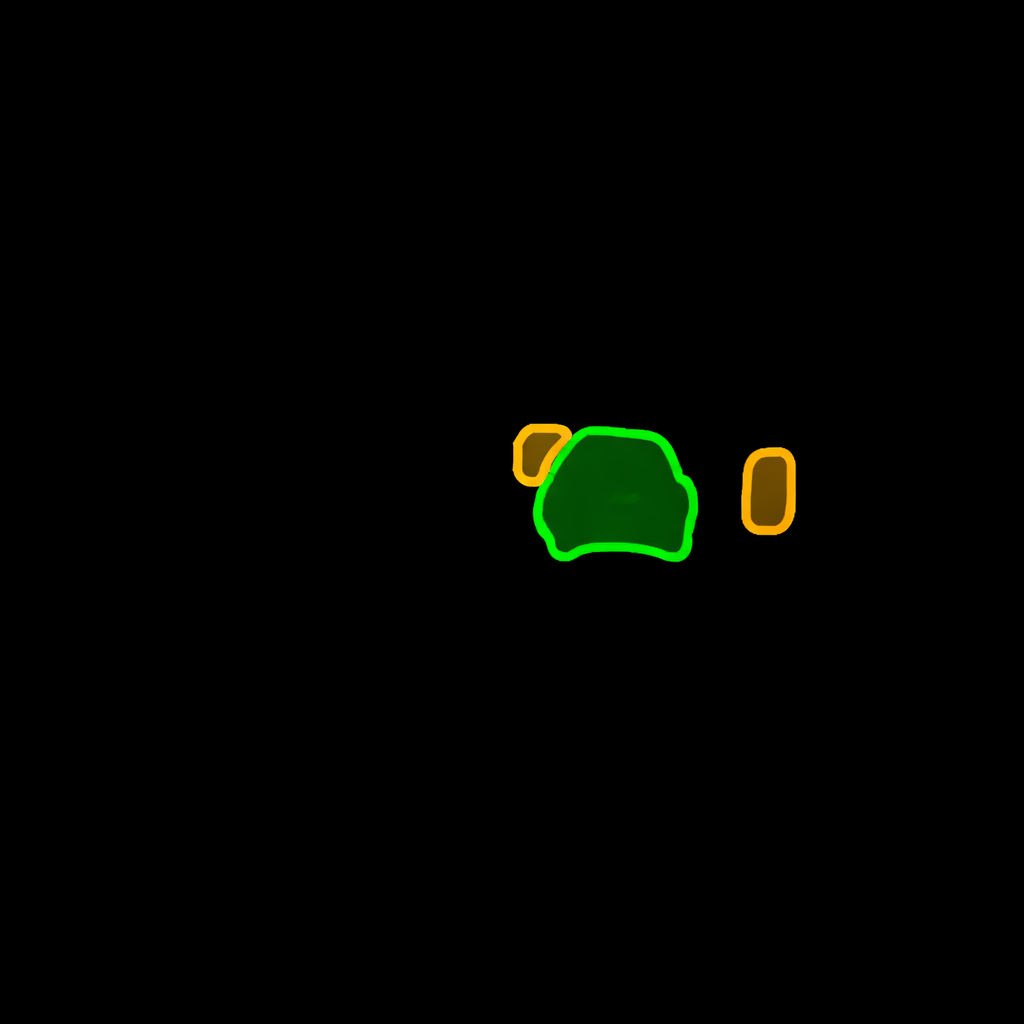} &
\includegraphics[width=\imgw,height=\imgh]{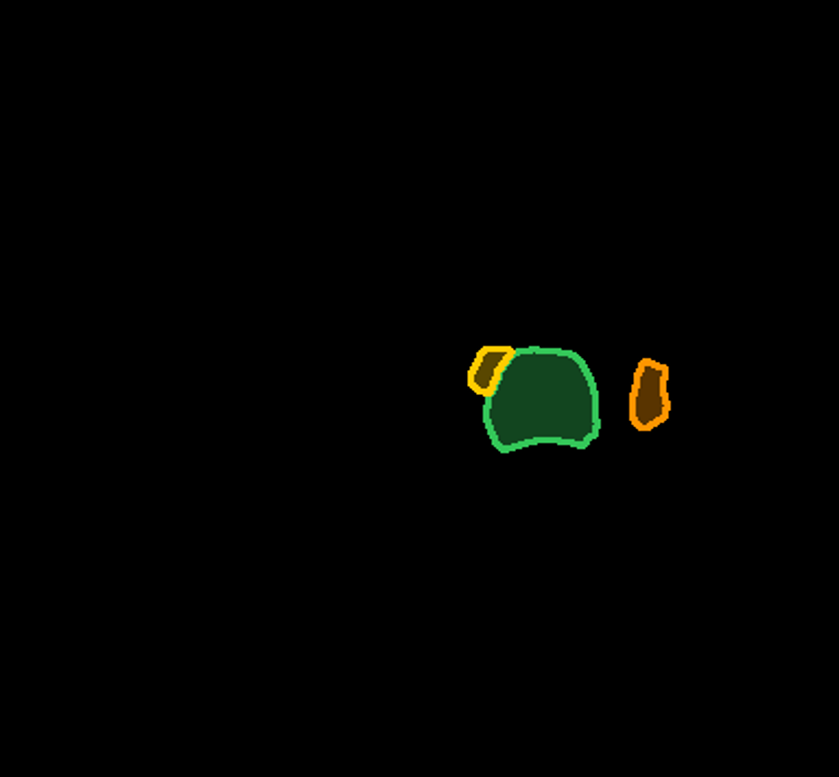} &
\includegraphics[width=\imgw,height=\imgh]{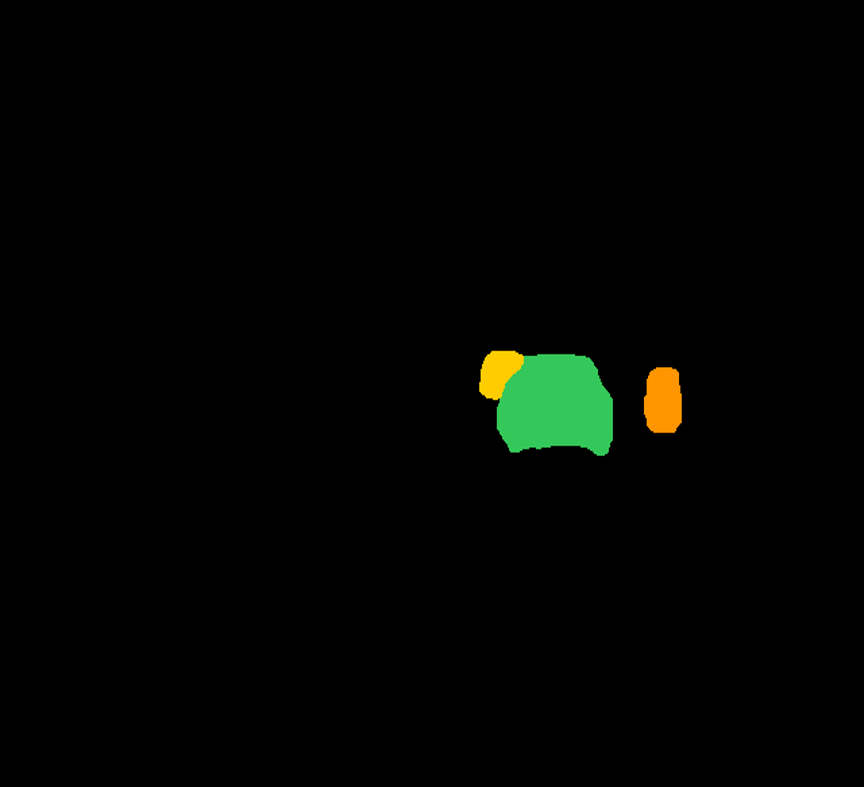} \\[-0.05em]

\includegraphics[width=\imgw,height=\imgh]{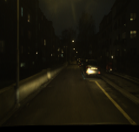} &
\includegraphics[width=\imgw,height=\imgh]{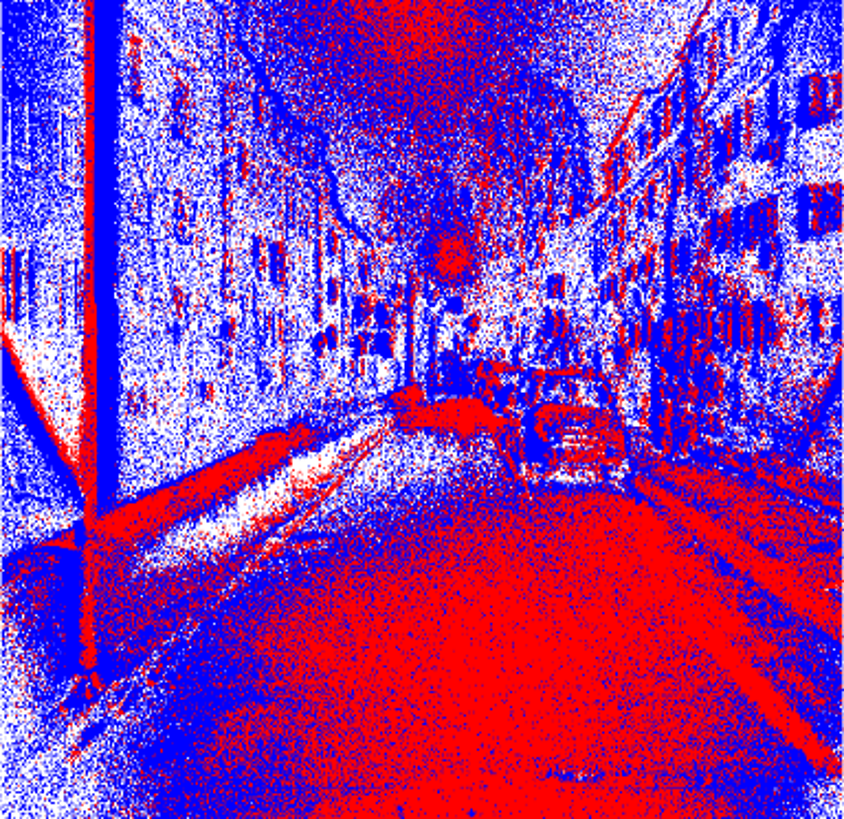} &
\includegraphics[width=\imgw,height=\imgh]{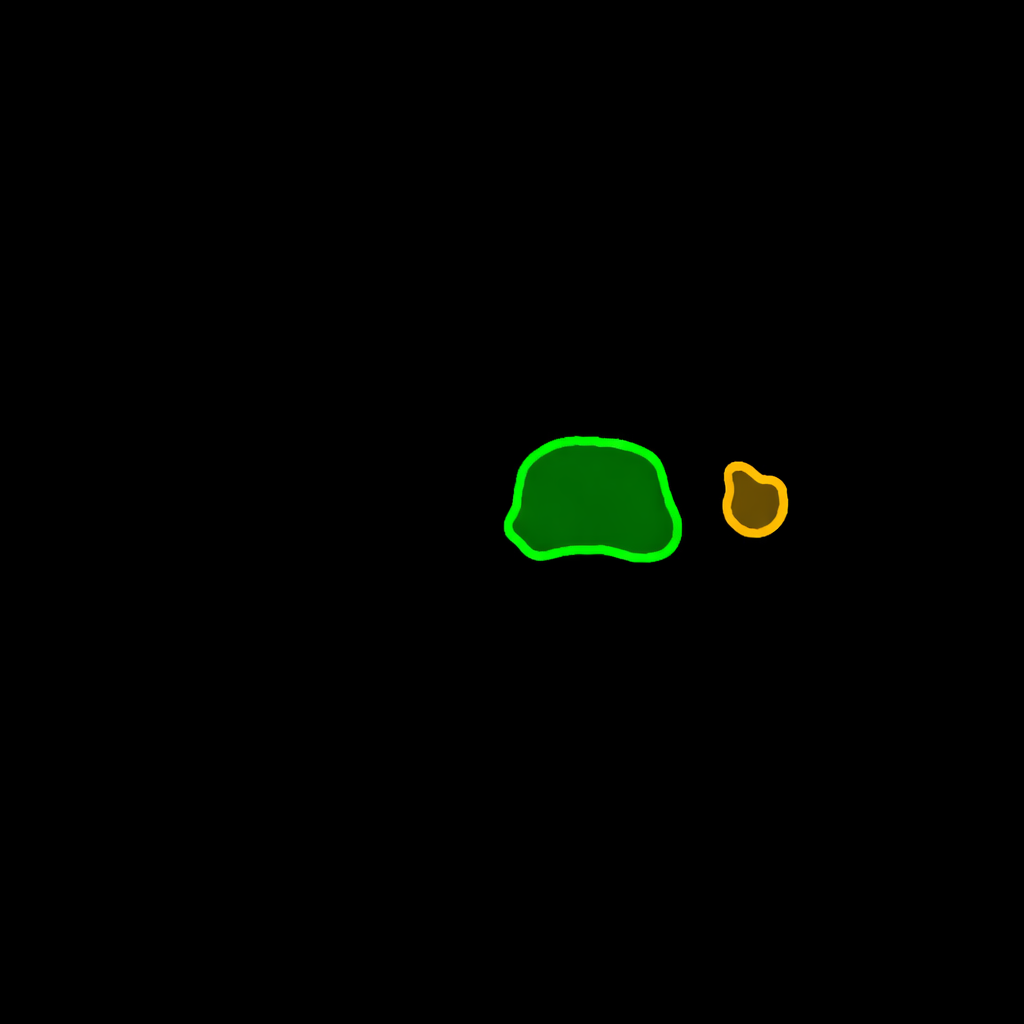} &
\includegraphics[width=\imgw,height=\imgh]{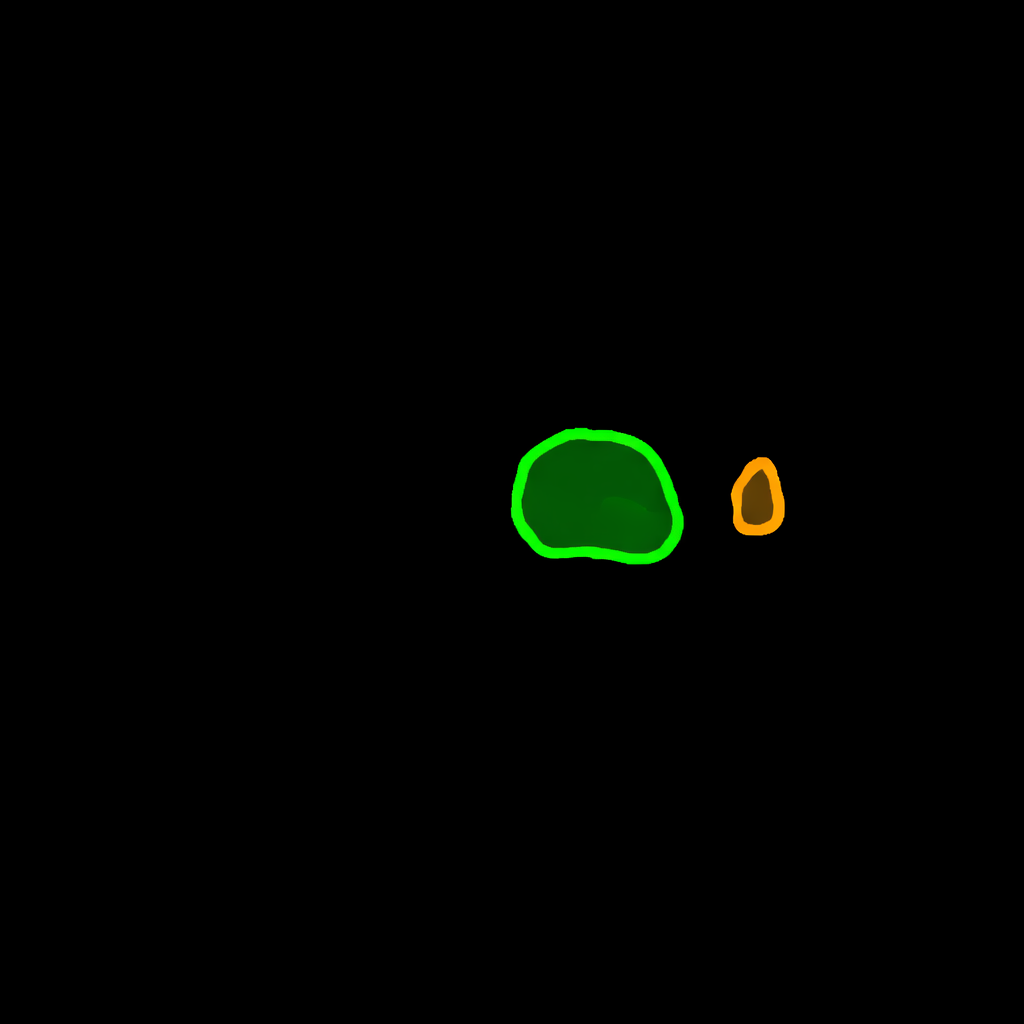} &
\includegraphics[width=\imgw,height=\imgh]{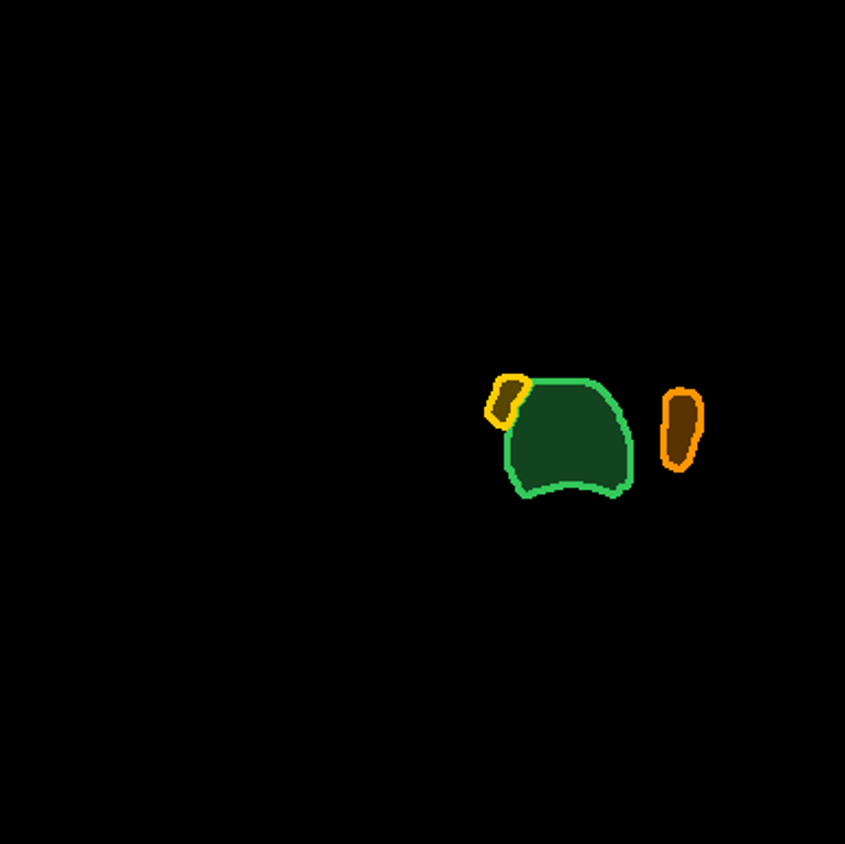} &
\includegraphics[width=\imgw,height=\imgh]{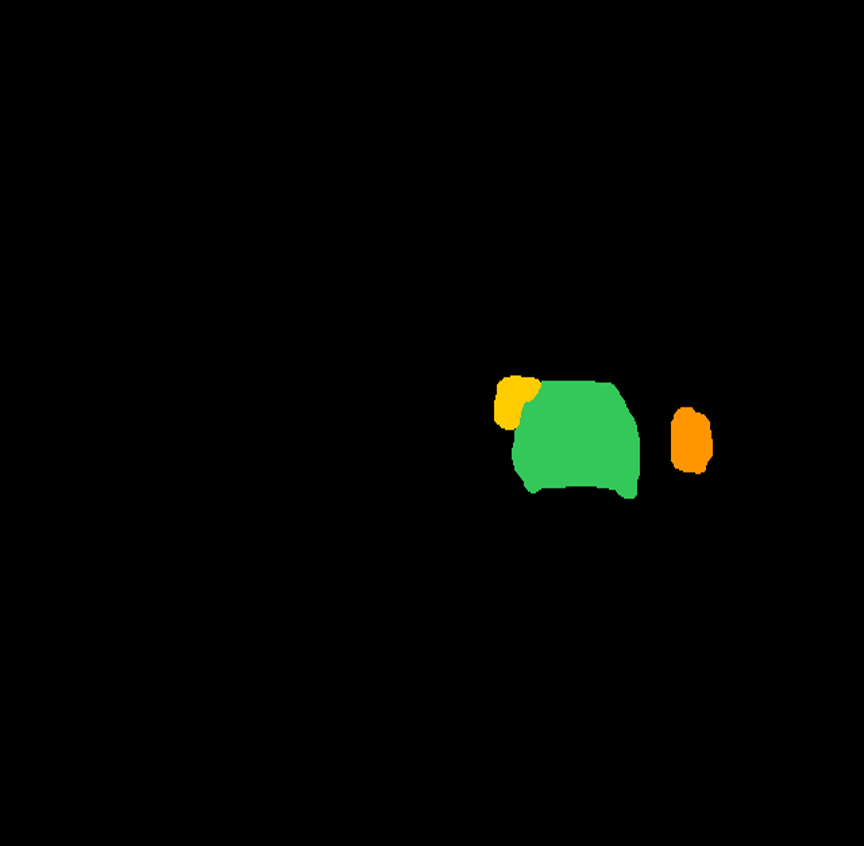} \\[-0.05em]

\includegraphics[width=\imgw,height=\imgh]{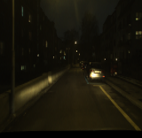} &
\includegraphics[width=\imgw,height=\imgh]{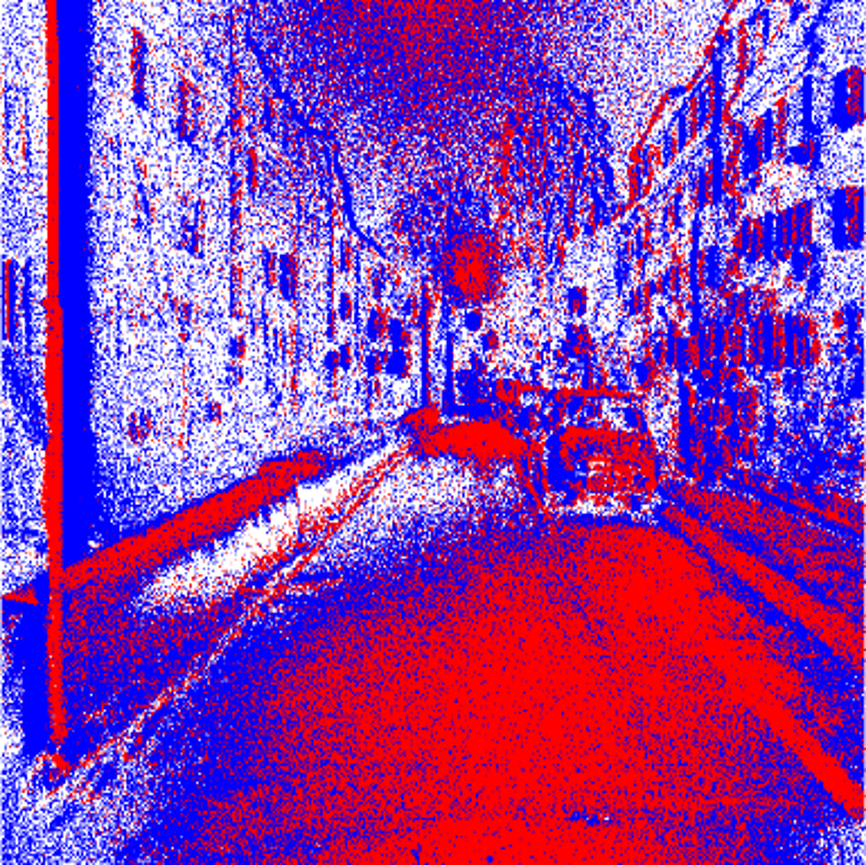} &
\includegraphics[width=\imgw,height=\imgh]{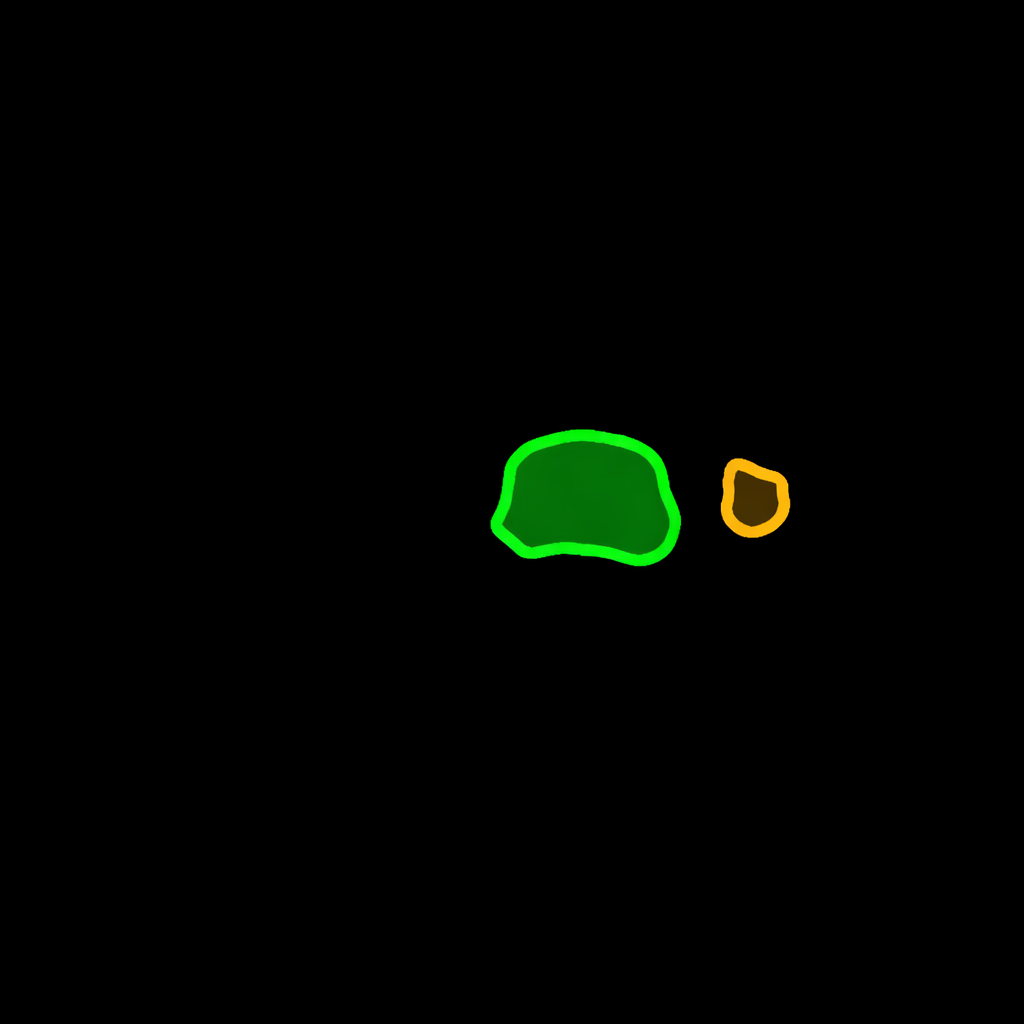} &
\includegraphics[width=\imgw,height=\imgh]{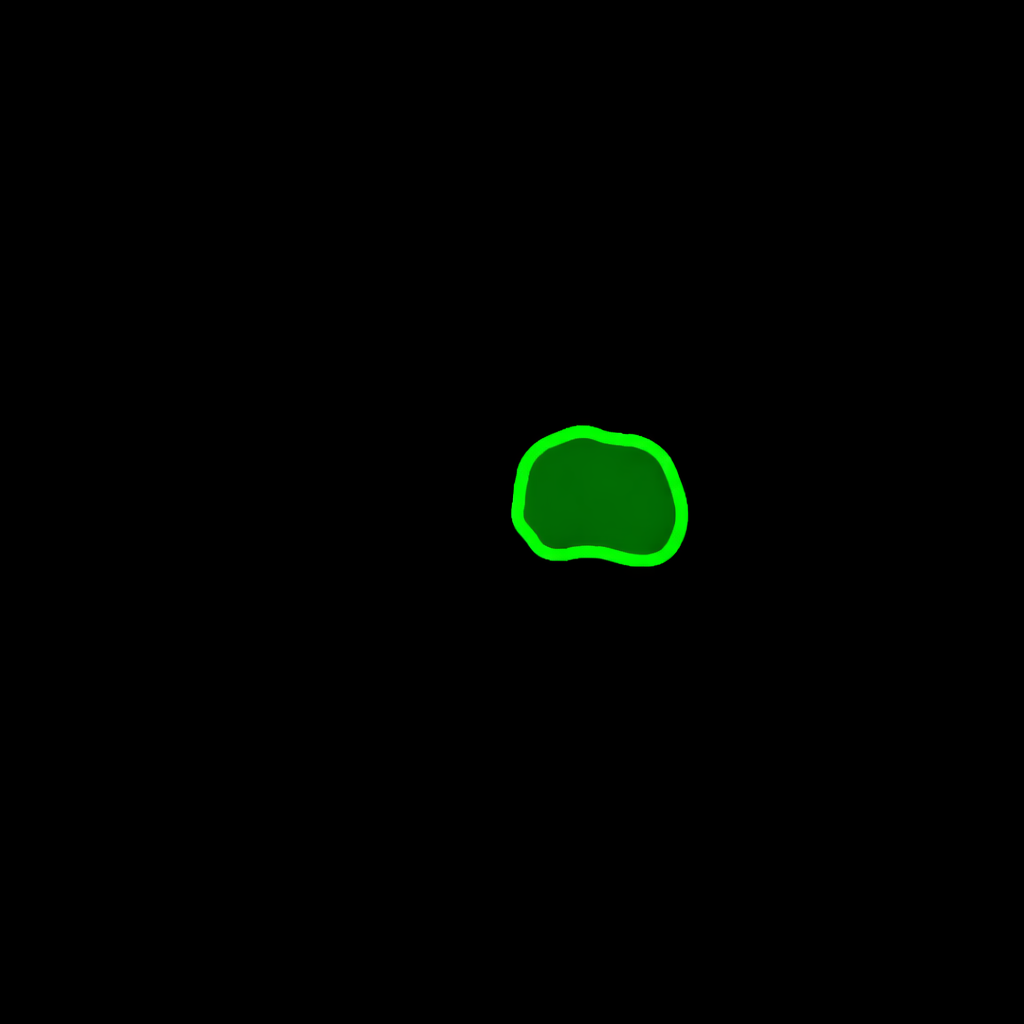} &
\includegraphics[width=\imgw,height=\imgh]{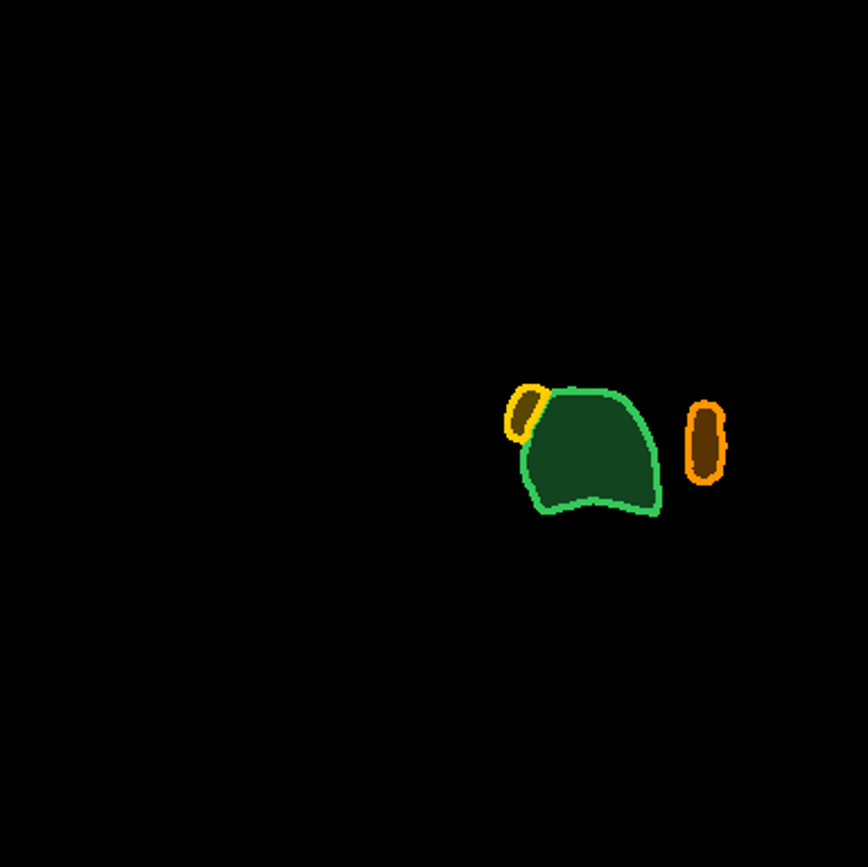} &
\includegraphics[width=\imgw,height=\imgh]{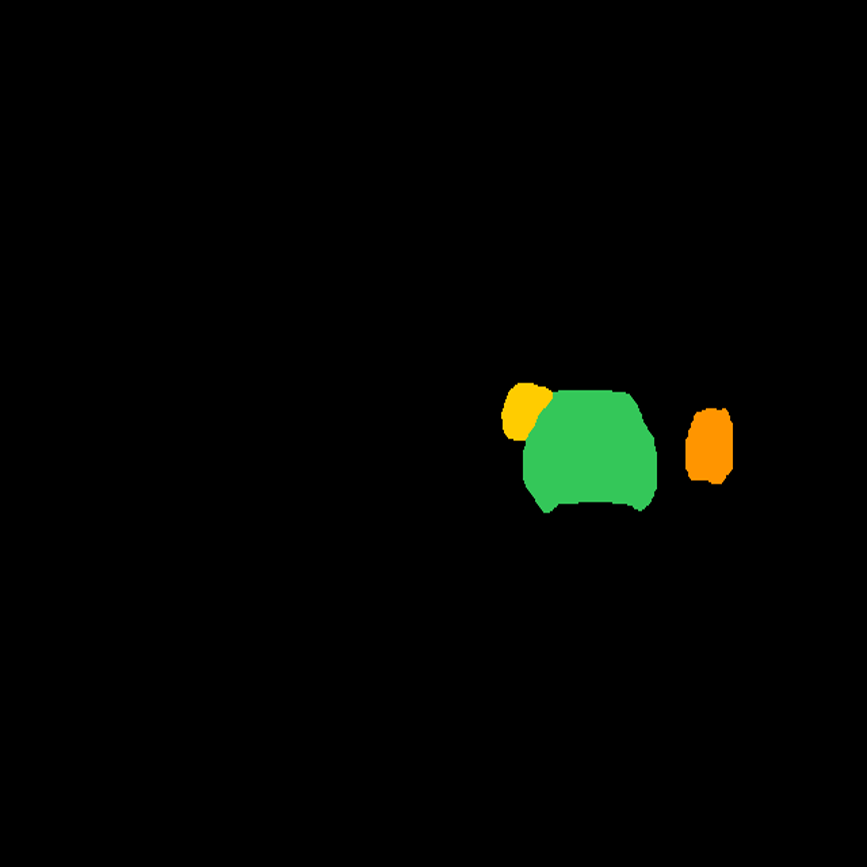} \\[0.25em]

{\scriptsize (a)} & {\scriptsize (b)} & {\scriptsize (c)} &
{\scriptsize (d)} & {\scriptsize (e)} & {\scriptsize (f)}
\end{tabular}

\vspace{-0.4em}
{\small
\caption{Qualitative comparison of promptable VOS under low-light conditions. Rows depict temporal progression. Columns: (a) RGB frames; (b) event-frame visualization rendered from the event voxel grid; (c) SAM2 on RGB; (d) SAM2 on E2VID-reconstructed frames; (e) E-MAPS on the event voxel grid; (f) ground-truth.}
\label{fig:samples}
}
\vspace{-1.2em}
\end{figure*}

\noindent\textbf{Zero-shot visual object tracking.}
To assess generalization beyond EViM, we evaluate E-MAPS without additional training on FE108~\cite{fe108} and VisEvent.
Following~\cite{samurai,dam4sam}, tracking is initialized with a bounding box, and predicted masks are converted into tight bounding boxes.
On FE108, event-only E-MAPS achieves 70.15 SR, outperforming EventSAM+ECOS by 4.90 points while remaining competitive with RGB and RGB--event methods.
On VisEvent, it performs particularly well under challenging conditions, ranking first in low illumination (LI) and surpassing all RGB-only methods under fast motion and motion blur (FM+MB).
We further explore RGB--event fusion by averaging aligned TinySAM RGB and adapted event features~\cite{zhang2026advancingcomplexvideoobject}.
Despite its simplicity, this variant outperforms dedicated RGB--event methods under low illumination, occlusion (FOC+POC), and background motion and clutter (BOM+BC), while remaining competitive on FE108 and FM+MB.
These findings demonstrate that representations learned for ePVS transfer effectively to conventional visual object tracking.

\noindent\textbf{Zero-shot interactive tracking on VisEvent.}
We further evaluate how E-MAPS responds to user corrections without additional training.
The protocol allows up to three online box corrections whenever the predicted IoU falls below 0.5.
While SAM2 and SAM-I2V operate on E2VID-reconstructed frames, E-MAPS processes events directly.
We measure the Success Rate before and after correction ($\mathrm{SR}_0$ and $\mathrm{SR}_3$), with relative improvement defined as $\Delta_{\mathrm{rel}}=(\mathrm{SR}_3-\mathrm{SR}_0)/\mathrm{SR}_0$.
E-MAPS achieves relative gains of 15.3\% overall and 30.8\% under NMO conditions, substantially exceeding the improvements of both baselines.
These results highlight its ability to make effective use of sparse user corrections, particularly in challenging tracking scenarios.

\begin{table}[t]
\centering
\vspace{-0.5em}
\caption{
\textbf{Controlled memory-interface ablation.}
Latency is measured on an RTX~3080.
}
\label{tab:ablation_mem_interface}
\vspace{-0.7em}
\scriptsize
\setlength{\tabcolsep}{3.2pt}
\renewcommand{\arraystretch}{0.88}

\begin{tabular}{@{}lcccc@{}}
\toprule
\rowcolor{Gray}
Injection path & EViM500 & EViM1000 & EViM-HM & Latency (ms) \\
\midrule

Past object feature fusion
& 63.25
& 61.18
& 56.47
& 26.04 \\

Decoder memory cross-attn.
& 82.61
& 79.51
& 74.86
& 152.00 \\

MAP w/o Sparse-Motion Adj.
& 83.10
& 78.51
& 74.31
& 33.46 \\

MAP + dense memory cross-attn.
& 83.02
& 79.54
& 75.68
& 158.91 \\

\rowcolor{gaincolor}
MAP (Ours)
& \textbf{84.38}
& \textbf{80.37}
& \textbf{76.41}
& 36.55 \\

\bottomrule
\end{tabular}

\end{table}

\subsection{Ablation Study}

\label{sec:ablations}
Table~\ref{tab:ablation_mem_interface} evaluates memory retrieval and decoder injection while keeping the event-derived scene and object memory fixed. Dense memory cross-attention, following SAM2.1 Tiny, achieves decent results at 152.00\,ms. In contrast, MAP improves performance while reducing latency to 36.55\,ms ($4.16\times$ faster). Combining MAP with dense cross-attention yields no further improvement. Removing Sparse-Motion Adjustment reduces performance, with the largest drop on EViM-HM. Tables~\ref{tab:ablation_1} and~\ref{tab:ablation_stage2} assess the training objectives. In Stage~I, the full objective improves $\mathcal{J}\&\mathcal{F}$ from 53.17 to 57.07 over Dice+L1 alone, translating to a downstream Stage-II improvement from 76.34 to 81.37. Removing Dice or L1 reduces the final score to 75.96 and 79.45, respectively, highlighting their complementary contributions. In Stage~II, ablating Dice, focal, occlusion, or IoU supervision lowers performance. Finally, Table~\ref{tab:ablation_map_ctx} examines memory token capacity and current-window conditioning early in the training stage. Increasing $K$ from 1 to 2 improves performance, while conditioning on current event evidence benefits both token budgets. Combining the two yields the best results, emphasizing the value of grounding compact memory representations in current observations.

\noindent
\begin{minipage}{\columnwidth}

\begin{minipage}[t]{0.43\linewidth}
    \vspace{0pt}
    \centering
    \captionof{table}{Stage-I loss ablation and downstream effect.}
    \label{tab:ablation_1}
    \vspace{-0.6em}
    \setlength{\tabcolsep}{1.2pt}
    \renewcommand{\arraystretch}{0.82}
    \scriptsize
    \begin{tabular}{lcc}
        \toprule
        \rowcolor{Gray}
        Setting & S-I & S-II \\
        \midrule
        w/o CORAL+InfoNCE & 53.17 & 76.34 \\
        w/o InfoNCE       & 53.81 & 75.89 \\
        w/o CORAL         & 52.99 & 69.78 \\
        w/o Dice          & 55.02 & 75.96 \\
        w/o L1            & 55.81 & 79.45 \\
        \rowcolor{gaincolor}
        Full & \textbf{57.07} & \textbf{81.37} \\
        \bottomrule
    \end{tabular}
\end{minipage}%
\hfill
\begin{minipage}[t]{0.29\linewidth}
    \vspace{0pt}
    \centering
    \captionof{table}{Stage-II objective ablation.}
    \label{tab:ablation_stage2}
    \vspace{-0.6em}
    \setlength{\tabcolsep}{1.5pt}
    \renewcommand{\arraystretch}{0.82}
    \scriptsize
    \begin{tabular}{lc}
        \toprule
        \rowcolor{Gray}
        Setting & $J\&F$ \\
        \midrule
        w/o Dice  & 56.78 \\
        w/o Focal & 67.98 \\
        w/o Occ.  & 77.87 \\
        w/o IoU   & 76.09 \\
        \rowcolor{gaincolor}
        Full  & \textbf{81.37} \\
        \bottomrule
    \end{tabular}
\end{minipage}%
\hfill
\begin{minipage}[t]{0.24\linewidth}
    \vspace{0pt}
    \centering
    \captionof{table}{MAP ablation.}
    \label{tab:ablation_map_ctx}
    \vspace{-0.6em}
    \setlength{\tabcolsep}{1.5pt}
    \renewcommand{\arraystretch}{0.82}
    \scriptsize
    \begin{tabular}{ccc}
        \toprule
        \rowcolor{Gray}
        $K$ & Cur. & $J\&F$ \\
        \midrule
        1 & $\times$   & 52.1 \\
        1 & \checkmark & 53.4 \\
        2 & $\times$   & 52.8 \\
        \rowcolor{gaincolor}
        2 & \checkmark & \textbf{54.6} \\
        \bottomrule
    \end{tabular}
\end{minipage}

\end{minipage}
\vspace{0.2em}

\section{Conclusion}
\noindent
We introduced ePVS, where arbitrary-window prompts initialize, correct, or introduce objects and predictions form identity-consistent masklets. We presented E-MAPS, an event-only framework that retains pretrained semantic and prompting priors while learning propagation and correction directly from event masklets. E2T aligns recurrent event features with the pretrained representation space, while MAP converts object-conditioned history into sparse prompts, allowing memory and user inputs to share one efficient decoder pathway. We also introduced EViM for scalable representation transfer and event-video learning. We extensively evaluate E-MAPS across prompt-once and interactive ePVS, and zero-shot tracking on FE108 and VisEvent. These evaluations show that E-MAPS provides accurate and efficient segmentation from sparse event streams.

\section*{Acknowledgments}
This research was funded by Khalifa University
of Science and Technology through the Faculty Start-Ups under Project ID: KU-INT-FSU-2005-8474000775. This research was also supported by KU Research Center for Advanced Intelligent Systems (AIS), Khalifa University of Science and Technology (KU-AIS).

{
    \small
    \bibliographystyle{ieeenat_fullname}
    \bibliography{main}

@String(CVPR= {IEEE Conf. Comput. Vis. Pattern Recog.})

@String(ICCV= {Int. Conf. Comput. Vis.})

@String(ECCV= {Eur. Conf. Comput. Vis.})

@String(NIPS= {Adv. Neural Inform. Process. Syst.})

@String(AAAI = {AAAI})

@String(CVPR  = {CVPR})

@String(ICCV  = {ICCV})

@String(ECCV  = {ECCV})

@String(NIPS  = {NeurIPS})

@String(CVPR  = {IEEE Conf. Comput. Vis. Pattern Recog.})

@String(ICCV  = {Int. Conf. Comput. Vis.})

@String(ECCV  = {Eur. Conf. Comput. Vis.})

@String(AAAI  = {AAAI})

@InProceedings{sam1,
    author    = {Kirillov, Alexander and Mintun, Eric and Ravi, Nikhila and Mao, Hanzi and Rolland, Chloe and Gustafson, Laura and Xiao, Tete and Whitehead, Spencer and Berg, Alexander C. and Lo, Wan-Yen and Dollar, Piotr and Girshick, Ross},
    title     = {Segment Anything},
    booktitle = {Proceedings of the IEEE/CVF International Conference on Computer Vision (ICCV)},
    month     = {October},
    year      = {2023},
    pages     = {4015-4026}
}

@inproceedings{sam2,
title={{SAM} 2: Segment Anything in Images and Videos},
author={Nikhila Ravi and Valentin Gabeur and Yuan-Ting Hu and Ronghang Hu and Chaitanya Ryali and Tengyu Ma and Haitham Khedr and Roman R{\"a}dle and Chloe Rolland and Laura Gustafson and Eric Mintun and Junting Pan and Kalyan Vasudev Alwala and Nicolas Carion and Chao-Yuan Wu and Ross Girshick and Piotr Dollar and Christoph Feichtenhofer},
booktitle={The Thirteenth International Conference on Learning Representations},
year={2025},
url={https://openreview.net/forum?id=Ha6RTeWMd0}
}

@inproceedings{sam3,
title={{SAM} 3: Segment Anything with Concepts},
author={Nicolas Carion and Laura Gustafson and Yuan-Ting Hu and Shoubhik Debnath and Ronghang Hu and Didac Suris Coll-Vinent and Chaitanya Ryali and Kalyan Vasudev Alwala and Haitham Khedr and Andrew Huang and Jie Lei and Tengyu Ma and Baishan Guo and Arpit Kalla and Markus Marks and Joseph Greer and Meng Wang and Peize Sun and Roman R{\"a}dle and Triantafyllos Afouras and Effrosyni Mavroudi and Katherine Xu and Tsung-Han Wu and Yu Zhou and Liliane Momeni and RISHI HAZRA and Shuangrui Ding and Sagar Vaze and Francois Porcher and Feng Li and Siyuan Li and Aishwarya Kamath and Ho Kei Cheng and Piotr Dollar and Nikhila Ravi and Kate Saenko and Pengchuan Zhang and Christoph Feichtenhofer},
booktitle={The Fourteenth International Conference on Learning Representations},
year={2026},
url={https://openreview.net/forum?id=r35clVtGzw}
}

@article{tinysam, 
title={TinySAM: Pushing the Envelope for Efficient Segment Anything Model}, 
volume={39}, 
url={https://ojs.aaai.org/index.php/AAAI/article/view/34255}, 
DOI={10.1609/aaai.v39i19.34255}, 
number={19}, 
journal={Proceedings of the AAAI Conference on Artificial Intelligence}, 
author={Shu, Han and Li, Wenshuo and Tang, Yehui and Zhang, Yiman and Chen, Yihao and Li, Houqiang and Wang, Yunhe and Chen, Xinghao}, 
year={2025}, 
month={Apr.}, 
pages={20470-20478}
}

@ARTICLE{ecos,
  author={Zhu, Lin and Chen, Xianzhang and Wang, Lizhi and Wang, Xiao and Tian, Yonghong and Huang, Hua},
  journal={IEEE Transactions on Pattern Analysis and Machine Intelligence}, 
  title={Continuous-Time Object Segmentation Using High Temporal Resolution Event Camera}, 
  year={2025},
  volume={47},
  number={2},
  pages={807-824},
  doi={10.1109/TPAMI.2024.3477591}}

@ARTICLE{dsec1,
  author={Gehrig, Mathias and Aarents, Willem and Gehrig, Daniel and Scaramuzza, Davide},
  journal={IEEE Robotics and Automation Letters}, 
  title={DSEC: A Stereo Event Camera Dataset for Driving Scenarios}, 
  year={2021},
  volume={6},
  number={3},
  pages={4947-4954},
  doi={10.1109/LRA.2021.3068942}}

@article{dsec2,
  title={Low-latency automotive vision with event cameras},
  author={Gehrig, Daniel and Scaramuzza, Davide},
  journal={Nature},
  volume={629},
  number={8014},
  pages={1034--1040},
  year={2024},
  publisher={Nature Publishing Group UK London}
}

@InProceedings{nimagenet,
    author    = {Kim, Junho and Bae, Jaehyeok and Park, Gangin and Zhang, Dongsu and Kim, Young Min},
    title     = {N-ImageNet: Towards Robust, Fine-Grained Object Recognition With Event Cameras},
    booktitle = {Proceedings of the IEEE/CVF International Conference on Computer Vision (ICCV)},
    month     = {October},
    year      = {2021},
    pages     = {2146-2156}
}

@article{imagenet,
  title={Imagenet large scale visual recognition challenge},
  author={Russakovsky, Olga and Deng, Jia and Su, Hao and Krause, Jonathan and Satheesh, Sanjeev and Ma, Sean and Huang, Zhiheng and Karpathy, Andrej and Khosla, Aditya and Bernstein, Michael and others},
  journal={International journal of computer vision},
  volume={115},
  number={3},
  pages={211--252},
  year={2015},
  publisher={Springer}
}

@Inbook{coral,
author="Sun, Baochen
and Feng, Jiashi
and Saenko, Kate",
title="Correlation Alignment for Unsupervised Domain Adaptation",
bookTitle="Domain Adaptation in Computer Vision Applications",
year="2017",
publisher="Springer International Publishing",
address="Cham",
pages="153--171",
isbn="978-3-319-58347-1",
doi="10.1007/978-3-319-58347-1_8",
url="https://doi.org/10.1007/978-3-319-58347-1_8"
}

@ARTICLE{events_survey,
  author={Gallego, Guillermo and Delbrück, Tobi and Orchard, Garrick and Bartolozzi, Chiara and Taba, Brian and Censi, Andrea and Leutenegger, Stefan and Davison, Andrew J. and Conradt, Jörg and Daniilidis, Kostas and Scaramuzza, Davide},
  journal={IEEE Transactions on Pattern Analysis and Machine Intelligence}, 
  title={Event-Based Vision: A Survey}, 
  year={2022},
  volume={44},
  number={1},
  pages={154-180},
  doi={10.1109/TPAMI.2020.3008413}}

@InProceedings{sun2022ess,
author="Sun, Zhaoning
and Messikommer, Nico
and Gehrig, Daniel
and Scaramuzza, Davide",
editor="Avidan, Shai
and Brostow, Gabriel
and Ciss{\'e}, Moustapha
and Farinella, Giovanni Maria
and Hassner, Tal",
title="ESS: Learning Event-Based Semantic Segmentation from Still Images",
booktitle="Computer Vision -- ECCV 2022",
year="2022",
publisher="Springer Nature Switzerland",
address="Cham",
pages="341--357",
isbn="978-3-031-19830-4"
}

@InProceedings{eventsam,
    author    = {Chen, Zhiwen and Zhu, Zhiyu and Zhang, Yifan and Hou, Junhui and Shi, Guangming and Wu, Jinjian},
    title     = {Segment Any Event Streams via Weighted Adaptation of Pivotal Tokens},
    booktitle = {Proceedings of the IEEE/CVF Conference on Computer Vision and Pattern Recognition (CVPR)},
    month     = {June},
    year      = {2024},
    pages     = {3890-3900}
}

@inproceedings{seal,
title={Segment Any Events with Language},
author={Seungjun Lee and Gim Hee Lee},
booktitle={The Fourteenth International Conference on Learning Representations},
year={2026},
url={https://openreview.net/forum?id=zxito57J6x}
}

@InProceedings{stm,
author = {Oh, Seoung Wug and Lee, Joon-Young and Xu, Ning and Kim, Seon Joo},
title = {Video Object Segmentation Using Space-Time Memory Networks},
booktitle = {Proceedings of the IEEE/CVF International Conference on Computer Vision (ICCV)},
month = {October},
year = {2019}
}

@InProceedings{xmem,
author="Cheng, Ho Kei
and Schwing, Alexander G.",
editor="Avidan, Shai
and Brostow, Gabriel
and Ciss{\'e}, Moustapha
and Farinella, Giovanni Maria
and Hassner, Tal",
title="XMem: Long-Term Video Object Segmentation with an Atkinson-Shiffrin Memory Model",
booktitle="Computer Vision -- ECCV 2022",
year="2022",
publisher="Springer Nature Switzerland",
address="Cham",
pages="640--658",
isbn="978-3-031-19815-1"
}

@InProceedings{deepcoral,
author="Sun, Baochen
and Saenko, Kate",
editor="Hua, Gang
and J{\'e}gou, Herv{\'e}",
title="Deep CORAL: Correlation Alignment for Deep Domain Adaptation",
booktitle="Computer Vision -- ECCV 2016 Workshops",
year="2016",
publisher="Springer International Publishing",
address="Cham",
pages="443--450",
isbn="978-3-319-49409-8"
}

@article{cpc,
  title={Representation learning with contrastive predictive coding},
  author={Oord, Aaron van den and Li, Yazhe and Vinyals, Oriol},
  journal={arXiv preprint arXiv:1807.03748},
  year={2018}
}

@INPROCEEDINGS{event_sam_adapter,
  author={Yao, Bowen and Deng, Yongjian and Liu, Yuhan and Chen, Hao and Li, Youfu and Yang, Zhen},
  booktitle={2024 IEEE International Conference on Robotics and Automation (ICRA)}, 
  title={SAM-Event-Adapter: Adapting Segment Anything Model for Event-RGB Semantic Segmentation}, 
  year={2024},
  volume={},
  number={},
  pages={9093-9100},
  doi={10.1109/ICRA57147.2024.10611127}}

@article{memorysam,
  publtype={informal},
  author={Chenfei Liao and Xu Zheng and Yuanhuiyi Lyu and Haiwei Xue and Yihong Cao and Jiawen Wang and Kailun Yang and Xuming Hu},
  title={MemorySAM: Memorize Modalities and Semantics with Segment Anything Model 2 for Multi-modal Semantic Segmentation},
  year={2025},
  month={March},
  cdate={1740787200000},
  journal={CoRR},
  volume={abs/2503.06700},
  url={https://doi.org/10.48550/arXiv.2503.06700}
}

@InProceedings{event_lowlight_vos,
    author    = {Li, Hebei and Wang, Jin and Yuan, Jiahui and Li, Yue and Weng, Wenming and Peng, Yansong and Zhang, Yueyi and Xiong, Zhiwei and Sun, Xiaoyan},
    title     = {Event-assisted Low-Light Video Object Segmentation},
    booktitle = {Proceedings of the IEEE/CVF Conference on Computer Vision and Pattern Recognition (CVPR)},
    month     = {June},
    year      = {2024},
    pages     = {3250-3259}
}

@InProceedings{sami2v,
    author    = {Mei, Haiyang and Zhang, Pengyu and Shou, Mike Zheng},
    title     = {SAM-I2V: Upgrading SAM to Support Promptable Video Segmentation with Less than 0.2\% Training Cost},
    booktitle = {Proceedings of the IEEE/CVF Conference on Computer Vision and Pattern Recognition (CVPR)},
    month     = {June},
    year      = {2025},
    pages     = {3417-3426}
}

@ARTICLE{vs,
  author={Zhou, Tianfei and Porikli, Fatih and Crandall, David J. and Van Gool, Luc and Wang, Wenguan},
  journal={IEEE Transactions on Pattern Analysis and Machine Intelligence}, 
  title={A Survey on Deep Learning Technique for Video Segmentation}, 
  year={2023},
  volume={45},
  number={6},
  pages={7099-7122},
  doi={10.1109/TPAMI.2022.3225573}}

@InProceedings{minima,
    author    = {Ren, Jiangwei and Jiang, Xingyu and Li, Zizhuo and Liang, Dingkang and Zhou, Xin and Bai, Xiang},
    title     = {MINIMA: Modality Invariant Image Matching},
    booktitle = {Proceedings of the IEEE/CVF Conference on Computer Vision and Pattern Recognition (CVPR)},
    month     = {June},
    year      = {2025},
    pages     = {23059-23068}
}

@InProceedings{roma,
    author    = {Edstedt, Johan and Sun, Qiyu and B\"okman, Georg and Wadenb\"ack, M\r{a}rten and Felsberg, Michael},
    title     = {RoMa: Robust Dense Feature Matching},
    booktitle = {Proceedings of the IEEE/CVF Conference on Computer Vision and Pattern Recognition (CVPR)},
    month     = {June},
    year      = {2024},
    pages     = {19790-19800}
}

@ARTICLE{e2vid,
  author={Rebecq, Henri and Ranftl, René and Koltun, Vladlen and Scaramuzza, Davide},
  journal={IEEE Transactions on Pattern Analysis and Machine Intelligence}, 
  title={High Speed and High Dynamic Range Video with an Event Camera}, 
  year={2021},
  volume={43},
  number={6},
  pages={1964-1980},
  doi={10.1109/TPAMI.2019.2963386}}

@InProceedings{bekuzarov2023xmem,
    author    = {Bekuzarov, Maksym and Bermudez, Ariana and Lee, Joon-Young and Li, Hao},
    title     = {XMem++: Production-level Video Segmentation From Few Annotated Frames},
    booktitle = {Proceedings of the IEEE/CVF International Conference on Computer Vision (ICCV)},
    month     = {October},
    year      = {2023},
    pages     = {635-644}
}

@InProceedings{Zhu_2019_CVPR,
author = {Zhu, Alex Zihao and Yuan, Liangzhe and Chaney, Kenneth and Daniilidis, Kostas},
title = {Unsupervised Event-Based Learning of Optical Flow, Depth, and Egomotion},
booktitle = {Proceedings of the IEEE/CVF Conference on Computer Vision and Pattern Recognition (CVPR)},
month = {June},
year = {2019}
}

@ARTICLE{vos1,
  title={Deep learning for video object segmentation: a review},
  author={Gao, Mingqi and Zheng, Feng and Yu, James JQ and Shan, Caifeng and Ding, Guiguang and Han, Jungong},
  journal={Artificial Intelligence Review},
  volume={56},
  number={1},
  pages={457--531},
  year={2023},
  publisher={Springer}
}

@InProceedings{osvos,
author = {Caelles, Sergi and Maninis, Kevis-Kokitsi and Pont-Tuset, Jordi and Leal-Taixe, Laura and Cremers, Daniel and Van Gool, Luc},
title = {One-Shot Video Object Segmentation},
booktitle = {Proceedings of the IEEE Conference on Computer Vision and Pattern Recognition (CVPR)},
month = {July},
year = {2017}
}

@INPROCEEDINGS{masktrack,
  author={Perazzi, Federico and Khoreva, Anna and Benenson, Rodrigo and Schiele, Bernt and Sorkine-Hornung, Alexander},
  booktitle={2017 IEEE Conference on Computer Vision and Pattern Recognition (CVPR)}, 
  title={Learning Video Object Segmentation from Static Images}, 
  year={2017},
  volume={},
  number={},
  pages={3491-3500},
  doi={10.1109/CVPR.2017.372}}

@INPROCEEDINGS{cutie,
  author={Cheng, Ho Kei and Oh, Seoung Wug and Price, Brian and Lee, Joon-Young and Schwing, Alexander},
  booktitle={2024 IEEE/CVF Conference on Computer Vision and Pattern Recognition (CVPR)}, 
  title={Putting the Object Back into Video Object Segmentation}, 
  year={2024},
  volume={},
  number={},
  pages={3151-3161},
  doi={10.1109/CVPR52733.2024.00304}}

@InProceedings{rmem,
    author    = {Zhou, Junbao and Pang, Ziqi and Wang, Yu-Xiong},
    title     = {RMem: Restricted Memory Banks Improve Video Object Segmentation},
    booktitle = {Proceedings of the IEEE/CVF Conference on Computer Vision and Pattern Recognition (CVPR)},
    month     = {June},
    year      = {2024},
    pages     = {18602-18611}
}

@inproceedings{aot,
 author = {Yang, Zongxin and Wei, Yunchao and Yang, Yi},
 booktitle = {Advances in Neural Information Processing Systems},
 editor = {M. Ranzato and A. Beygelzimer and Y. Dauphin and P.S. Liang and J. Wortman Vaughan},
 pages = {2491--2502},
 publisher = {Curran Associates, Inc.},
 title = {Associating Objects with Transformers for Video Object Segmentation},
 url = {https://proceedings.neurips.cc/paper_files/paper/2021/file/147702db07145348245dc5a2f2fe5683-Paper.pdf},
 volume = {34},
 year = {2021}
}

@inproceedings{stcn,
 author = {Cheng, Ho Kei and Tai, Yu-Wing and Tang, Chi-Keung},
 booktitle = {Advances in Neural Information Processing Systems},
 editor = {M. Ranzato and A. Beygelzimer and Y. Dauphin and P.S. Liang and J. Wortman Vaughan},
 pages = {11781--11794},
 publisher = {Curran Associates, Inc.},
 title = {Rethinking Space-Time Networks with Improved Memory Coverage for Efficient Video Object Segmentation},
 url = {https://proceedings.neurips.cc/paper_files/paper/2021/file/61b4a64be663682e8cb037d9719ad8cd-Paper.pdf},
 volume = {34},
 year = {2021}
}

@ARTICLE{lim1,
  author={Sun, Mingjie and Xiao, Jimin and Lim, Eng Gee and Zhao, Cairong and Zhao, Yao},
  journal={IEEE Transactions on Circuits and Systems for Video Technology}, 
  title={Unified Multi-Modality Video Object Segmentation Using Reinforcement Learning}, 
  year={2024},
  volume={34},
  number={8},
  pages={6722-6734},
  doi={10.1109/TCSVT.2023.3284165}}

@InProceedings{events_survey_2,
author="Chakravarthi, Bharatesh
and Verma, Aayush Atul
and Daniilidis, Kostas
and Fermuller, Cornelia
and Yang, Yezhou",
editor="Del Bue, Alessio
and Canton, Cristian
and Pont-Tuset, Jordi
and Tommasi, Tatiana",
title="Recent Event Camera Innovations: A Survey",
booktitle="Computer Vision -- ECCV 2024 Workshops",
year="2025",
publisher="Springer Nature Switzerland",
address="Cham",
pages="342--376",
isbn="978-3-031-92460-6"
}

@article{events_survey_3,
  title={Deep learning for event-based vision: A comprehensive survey and benchmarks},
  author={Zheng, Xu and Liu, Yexin and Lu, Yunfan and Hua, Tongyan and Pan, Tianbo and Zhang, Weiming and Tao, Dacheng and Wang, Lin},
  journal={arXiv preprint arXiv:2302.08890},
  year={2023}
}

@article{evsam,
author = {Ding, Yi and Yao, Bowen and Liu, Yuhan and Chen, Hao and Ding, Ding and Yang, Zhen and Li, Youfu and Deng, Yongjian},
title = {EvSAM: Segment Anything Model with Event-based Assistance},
year = {2026},
publisher = {Association for Computing Machinery},
address = {New York, NY, USA},
issn = {1551-6857},
url = {https://doi.org/10.1145/3786794},
doi = {10.1145/3786794},
note = {Just Accepted},
journal = {ACM Trans. Multimedia Comput. Commun. Appl.},
month = jan
}

@ARTICLE{event_det,
  author={Li, Jianing and Li, Jia and Zhu, Lin and Xiang, Xijie and Huang, Tiejun and Tian, Yonghong},
  journal={IEEE Transactions on Image Processing}, 
  title={Asynchronous Spatio-Temporal Memory Network for Continuous Event-Based Object Detection}, 
  year={2022},
  volume={31},
  number={},
  pages={2975-2987},
  doi={10.1109/TIP.2022.3162962}}

@InProceedings{evsegnet,
author = {Alonso, Inigo and Murillo, Ana C.},
title = {EV-SegNet: Semantic Segmentation for Event-Based Cameras},
booktitle = {Proceedings of the IEEE/CVF Conference on Computer Vision and Pattern Recognition (CVPR) Workshops},
month = {June},
year = {2019}
}

@ARTICLE{eisnet,
  author={Xie, Bochen and Deng, Yongjian and Shao, Zhanpeng and Li, Youfu},
  journal={IEEE Transactions on Multimedia}, 
  title={EISNet: A Multi-Modal Fusion Network for Semantic Segmentation With Events and Images}, 
  year={2024},
  volume={26},
  number={},
  pages={8639-8650},
  doi={10.1109/TMM.2024.3380255}}

@ARTICLE{cmx,
  author={Zhang, Jiaming and Liu, Huayao and Yang, Kailun and Hu, Xinxin and Liu, Ruiping and Stiefelhagen, Rainer},
  journal={IEEE Transactions on Intelligent Transportation Systems}, 
  title={CMX: Cross-Modal Fusion for RGB-X Semantic Segmentation With Transformers}, 
  year={2023},
  volume={24},
  number={12},
  pages={14679-14694},
  doi={10.1109/TITS.2023.3300537}}

@InProceedings{cmnext,
    author    = {Zhang, Jiaming and Liu, Ruiping and Shi, Hao and Yang, Kailun and Rei{\ss}, Simon and Peng, Kunyu and Fu, Haodong and Wang, Kaiwei and Stiefelhagen, Rainer},
    title     = {Delivering Arbitrary-Modal Semantic Segmentation},
    booktitle = {Proceedings of the IEEE/CVF Conference on Computer Vision and Pattern Recognition (CVPR)},
    month     = {June},
    year      = {2023},
    pages     = {1136-1147}
}

@InProceedings{openess,
    author    = {Kong, Lingdong and Liu, Youquan and Ng, Lai Xing and Cottereau, Benoit R. and Ooi, Wei Tsang},
    title     = {OpenESS: Event-based Semantic Scene Understanding with Open Vocabularies},
    booktitle = {Proceedings of the IEEE/CVF Conference on Computer Vision and Pattern Recognition (CVPR)},
    month     = {June},
    year      = {2024},
    pages     = {15686-15698}
}

@article{deaot,
  title={Decoupling features in hierarchical propagation for video object segmentation},
  author={Yang, Zongxin and Yang, Yi},
  journal={Advances in Neural Information Processing Systems},
  volume={35},
  pages={36324--36336},
  year={2022}
}

@inproceedings{fe108,
  title={Object tracking by jointly exploiting frame and event domain},
  author={Zhang, Jiqing and Yang, Xin and Fu, Yingkai and Wei, Xiaopeng and Yin, Baocai and Dong, Bo},
  booktitle={Proceedings of the IEEE/CVF international conference on computer vision},
  pages={13043--13052},
  year={2021}
}

@misc{samurai,
      title={SAMURAI: Adapting Segment Anything Model for Zero-Shot Visual Tracking with Motion-Aware Memory}, 
      author={Cheng-Yen Yang and Hsiang-Wei Huang and Wenhao Chai and Zhongyu Jiang and Jenq-Neng Hwang},
      year={2024},
      eprint={2411.11922},
      archivePrefix={arXiv},
      primaryClass={cs.CV},
      url={https://arxiv.org/abs/2411.11922}, 
}

@InProceedings{dam4sam,
  author = {Videnovic, Jovana and Lukezic, Alan and Kristan, Matej},
  title = {A Distractor-Aware Memory for Visual Object Tracking with {SAM2}},
  booktitle = {Comp. Vis. Patt. Recognition},
  year = {2025}
}

@ARTICLE{visevent,
  author={Wang, Xiao and Li, Jianing and Zhu, Lin and Zhang, Zhipeng and Chen, Zhe and Li, Xin and Wang, Yaowei and Tian, Yonghong and Wu, Feng},
  journal={IEEE Transactions on Cybernetics}, 
  title={VisEvent: Reliable Object Tracking via Collaboration of Frame and Event Flows}, 
  year={2024},
  volume={54},
  number={3},
  pages={1997-2010},
  doi={10.1109/TCYB.2023.3318601}}

@misc{radford2021learningtransferablevisualmodels,
      title={Learning Transferable Visual Models From Natural Language Supervision}, 
      author={Alec Radford and Jong Wook Kim and Chris Hallacy and Aditya Ramesh and Gabriel Goh and Sandhini Agarwal and Girish Sastry and Amanda Askell and Pamela Mishkin and Jack Clark and Gretchen Krueger and Ilya Sutskever},
      year={2021},
      eprint={2103.00020},
      archivePrefix={arXiv},
      primaryClass={cs.CV},
      url={https://arxiv.org/abs/2103.00020}, 
}

@misc{dai2017deformableconvolutionalnetworks,
      title={Deformable Convolutional Networks}, 
      author={Jifeng Dai and Haozhi Qi and Yuwen Xiong and Yi Li and Guodong Zhang and Han Hu and Yichen Wei},
      year={2017},
      eprint={1703.06211},
      archivePrefix={arXiv},
      primaryClass={cs.CV},
      url={https://arxiv.org/abs/1703.06211}, 
}

@misc{glamm,
      title={GLaMM: Pixel Grounding Large Multimodal Model}, 
      author={Hanoona Rasheed and Muhammad Maaz and Sahal Shaji Mullappilly and Abdelrahman Shaker and Salman Khan and Hisham Cholakkal and Rao M. Anwer and Erix Xing and Ming-Hsuan Yang and Fahad S. Khan},
      year={2024},
      eprint={2311.03356},
      archivePrefix={arXiv},
      primaryClass={cs.CV},
      url={https://arxiv.org/abs/2311.03356}, 
}

@misc{videoglamm,
      title={VideoGLaMM: A Large Multimodal Model for Pixel-Level Visual Grounding in Videos}, 
      author={Shehan Munasinghe and Hanan Gani and Wenqi Zhu and Jiale Cao and Eric Xing and Fahad Shahbaz Khan and Salman Khan},
      year={2025},
      eprint={2411.04923},
      archivePrefix={arXiv},
      primaryClass={cs.CV},
      url={https://arxiv.org/abs/2411.04923}, 
}

@misc{videolisa,
      title={One Token to Seg Them All: Language Instructed Reasoning Segmentation in Videos}, 
      author={Zechen Bai and Tong He and Haiyang Mei and Pichao Wang and Ziteng Gao and Joya Chen and Lei Liu and Zheng Zhang and Mike Zheng Shou},
      year={2024},
      eprint={2409.19603},
      archivePrefix={arXiv},
      primaryClass={cs.CV},
      url={https://arxiv.org/abs/2409.19603}, 
}

@inbook{Hamann_2025,
   title={MouseSIS: A Frames-and-Events Dataset for Space-Time Instance Segmentation of Mice},
   ISBN={9783031924606},
   ISSN={1611-3349},
   url={http://dx.doi.org/10.1007/978-3-031-92460-6_10},
   DOI={10.1007/978-3-031-92460-6_10},
   booktitle={Computer Vision – ECCV 2024 Workshops},
   publisher={Springer Nature Switzerland},
   author={Hamann, Friedhelm and Li, Hanxiong and Mieske, Paul and Lewejohann, Lars and Gallego, Guillermo},
   year={2025},
   pages={156–173} }

@misc{li2025efficienteventbasedsemanticsegmentation,
      title={Efficient Event-Based Semantic Segmentation via Exploiting Frame-Event Fusion: A Hybrid Neural Network Approach}, 
      author={Hebei Li and Yansong Peng and Jiahui Yuan and Peixi Wu and Jin Wang and Yueyi Zhang and Xiaoyan Sun},
      year={2025},
      eprint={2507.03765},
      archivePrefix={arXiv},
      primaryClass={cs.CV},
      url={https://arxiv.org/abs/2507.03765}, 
}

@inproceedings{vaswani2023attentionneed,
author = {Vaswani, Ashish and Shazeer, Noam and Parmar, Niki and Uszkoreit, Jakob and Jones, Llion and Gomez, Aidan N. and Kaiser, {\L}ukasz and Polosukhin, Illia},
title = {Attention is all you need},
year = {2017},
isbn = {9781510860964},
publisher = {Curran Associates Inc.},
address = {Red Hook, NY, USA},
booktitle = {Proceedings of the 31st International Conference on Neural Information Processing Systems},
pages = {6000–6010},
numpages = {11},
location = {Long Beach, California, USA},
series = {NIPS'17}
}

@inproceedings{Yao_2025,
   title={Event-Guided Low-Light Video Semantic Segmentation},
   url={http://dx.doi.org/10.1109/WACV61041.2025.00329},
   DOI={10.1109/wacv61041.2025.00329},
   booktitle={2025 IEEE/CVF Winter Conference on Applications of Computer Vision (WACV)},
   publisher={IEEE},
   author={Yao, Zhen and Chuah, Mooi Choo},
   year={2025},
   month=Feb, pages={3330–3341} }

@article{chen2026scaleevent,
  title={Scaling Dense Event-Stream Pretraining from Visual Foundation Models},
  author={Chen, Zhiwen and Hou, Junhui and Zhu, Zhiyu and Wu, Jinjian and Shi, Guangming},
  journal={arXiv preprint arXiv:2603.03969},
  year={2026}
}

@article{polizzi2026realm,
  title={REALM: An RGB and Event Aligned Latent Manifold for Cross-Modal Perception},
  author={Polizzi, Vincenzo and Lindell, David B and Kelly, Jonathan},
  journal={arXiv preprint arXiv:2605.00271},
  year={2026}
}

@inproceedings{baek2025evolve,
  title={EVOLVE: Event-guided deformable feature transfer and dual-memory refinement for low-light video object segmentation},
  author={Baek, Jong-Hyeon and Oh, Jiwon and Koh, Yeong Jun},
  booktitle={2025 IEEE/CVF International Conference on Computer Vision (ICCV)},
  pages={11273--11282},
  year={2025},
  organization={IEEE}
}

@inproceedings{zhang2023frameevent,
  title={Frame-event alignment and fusion network for high frame rate tracking},
  author={Zhang, Jiqing and Wang, Yuanchen and Liu, Wenxi and Li, Meng and Bai, Jinpeng and Yin, Baocai and Yang, Xin},
  booktitle={2023 IEEE/CVF Conference on Computer Vision and Pattern Recognition (CVPR)},
  pages={9781--9790},
  year={2023},
  organization={IEEE}
}

@inproceedings{wang2024eventvot,
  title={Event stream-based visual object tracking: A high-resolution benchmark dataset and a novel baseline},
  author={Wang, Xiao and Wang, Shiao and Tang, Chuanming and Zhu, Lin and Jiang, Bo and Tian, Yonghong and Tang, Jin},
  booktitle={2024 IEEE/CVF Conference on Computer Vision and Pattern Recognition (CVPR)},
  pages={19248--19257},
  year={2024},
  organization={IEEE}
}

@misc{zhang2026advancingcomplexvideoobject,
      title={Advancing Complex Video Object Segmentation via Progressive Concept Construction}, 
      author={Zhixiong Zhang and Shuangrui Ding and Xiaoyi Dong and Songxin He and Jianfan Lin and Junsong Tang and Yuhang Zang and Yuhang Cao and Dahua Lin and Jiaqi Wang},
      year={2026},
      eprint={2507.15852},
      archivePrefix={arXiv},
      primaryClass={cs.CV},
      url={https://arxiv.org/abs/2507.15852}, 
}
}

\clearpage

\end{document}